\pdfoutput=1

\documentclass[11pt]{article}

\usepackage[final]{acl}

\usepackage[activate={true,nocompatibility},final,tracking=true,
kerning=true,spacing=true,factor=1100,stretch=10,shrink=10]{microtype}
\usepackage{color}
\usepackage{times}
\usepackage{latexsym}
\usepackage{amsmath}
\usepackage{balance}

\usepackage[T1]{fontenc}

\usepackage[utf8]{inputenc}
\usepackage{graphicx} 
\usepackage{tikz}

\usepackage{booktabs}
\usepackage{multirow}
\usepackage{multicol}
\usepackage{makecell}
\usepackage{tabularx}
\usepackage{nicematrix}
\usepackage{graphicx}
\usepackage{threeparttable}

\usetikzlibrary{trees,positioning,shapes,shadows,arrows.meta}

\usepackage[edges]{forest}
\definecolor{hidden-draw}{RGB}{205, 44, 36}
\definecolor{hidden-blue}{RGB}{194,232,247}
\definecolor{hidden-orange}{RGB}{243,202,120}
\definecolor{hidden-yellow}{RGB}{242,244,193}
\definecolor{tree-level-1}{RGB}{245,20,85}
\definecolor{tree-level-2}{RGB}{246,86,118}
\definecolor{tree-level-3}{RGB}{248,177,193}
\definecolor{tree-leaf}{RGB}{176,230,198}
\usepackage{microtype}
\usepackage{bbding}
\usepackage{inconsolata}

\definecolor{darkgreen}{RGB}{0,150,0} 
\newcommand{\upnum}[1]{\textcolor{darkgreen}{$\uparrow$\footnotesize{#1}}}
\newcommand{\downnum}[1]{\textcolor{blue}{$\downarrow$\footnotesize{#1}}}
\definecolor{bluegray}{HTML}{647D87}
\newcommand{\demonstrationfigure}[2]{
\begin{tikzpicture}[
    every node/.style={outer sep=0},
    window/.style={rectangle, draw=black, rounded corners, thick, align=left, font=\small, inner xsep=0pt, inner ysep=3mm},
    titlestyle/.style={text=white, font=\bfseries},
]
\node[window] (window) at (0,-0.2cm) [minimum width=\linewidth, text width=\linewidth-4mm, anchor=north] {#2};
\draw[draw=black, fill=bluegray, thick] (-\linewidth/2,-0.3cm)
    [sharp corners] -- (\linewidth/2, -0.3cm)
    [rounded corners=3pt] -- (\linewidth/2, 0.25cm)
    [rounded corners=3pt] -- (-\linewidth/2, 0.25cm)
    [sharp corners] -- cycle;
\node[titlestyle] (title) at (0,-0.3mm) {#1};
\end{tikzpicture}%
}

\title{CausalEval: Towards Better Causal Reasoning in Language Models}

\author{
    Longxuan Yu\thanks{\ \ Equal contribution.}\textsuperscript{$\dag$}, 
    Delin Chen\footnotemark[1]\textsuperscript{$\ddag$}, 
    Siheng Xiong\footnotemark[1]\textsuperscript{$+$}, 
    Qingyang Wu\textsuperscript{$\S$},\\
    \textbf{Qingzhen Liu}\textsuperscript{$\diamondsuit$},
    \textbf{Dawei Li}\textsuperscript{$\heartsuit$}, 
    \textbf{Zhikai Chen}\textsuperscript{$\clubsuit$}, 
    \textbf{Xiaoze Liu}\textsuperscript{$\spadesuit$}, 
    \textbf{Liangming Pan}\textsuperscript{$\diamondsuit$}\\
    \textsuperscript{$\dag$}UC San Diego \quad 
    \textsuperscript{$\ddag$}UMass Amherst \quad 
    \textsuperscript{$+$}Georgia Tech \quad
    \textsuperscript{$\S$}UCLA  \\
    \textsuperscript{$\heartsuit$}Arizona State University \quad 
    \textsuperscript{$\clubsuit$}Michigan State University \\
    \textsuperscript{$\spadesuit$}Purdue University \quad 
    \textsuperscript{$\diamondsuit$}University of Arizona\\
    \texttt{loy004@ucsd.edu} \quad
    \texttt{delinchen@umass.edu} \quad
\texttt{liangmingpan@arizona.edu}
}

\begin{document}
\maketitle
\begin{abstract}
Causal reasoning (CR) is a crucial aspect of intelligence, essential for problem-solving, decision-making, and understanding the world. While language models (LMs) can generate rationales for their outputs, their ability to reliably perform causal reasoning remains uncertain, often falling short in tasks requiring a deep understanding of causality. 
In this paper, we introduce CausalEval, a comprehensive review of research aimed at enhancing LMs for causal reasoning, coupled with an empirical evaluation of current models and methods.
We categorize existing methods based on the role of LMs: either as reasoning engines or as helpers providing knowledge or data to traditional CR methods, followed by a detailed discussion of methodologies in each category. 
We then assess the performance of current LMs and various enhancement methods on a range of causal reasoning tasks, providing key findings and in-depth analysis. 
Finally, we present insights from current studies and highlight promising directions for future research. 
We aim for this work to serve as a comprehensive resource, fostering further advancements in causal reasoning with LMs.\footnote{Resources are available at \url{https://github.com/chendl02/Awesome-LLM-causal-reasoning}.}
\end{abstract}

\section{Introduction}
\label{Introduction}

Large Language Models (LLMs) \cite{openai2023gpt4, dubey2024llama} have achieved significant success across various domains. Causal reasoning (CR), which includes causal discovery and inference, is a fundamental aspect of human intelligence, facilitating effective problem-solving, decision-making \cite{10.1145/3629169}, and understanding of the world \cite{yao2021survey}, with applications in healthcare \cite{mahmooda2014framingham, prosperi2020causal}, economics \cite{celli2022causal, vemuri2015causal}, biology \cite{ramsey2018fask}, among other fields. While LLMs can often generate rationales for their outputs \cite{huang2022largelanguagemodelsselfimprove}, it remains controversial whether they genuinely understand causal relationships \cite{kiciman2023causal}. For example, \citet{liu2024llmscapabledatabasedstatistical} found that LLMs struggle to integrate pre-trained causal knowledge in statistical reasoning tasks.

\begin{figure}[t]
 \centering
\includegraphics[width=0.485\textwidth]{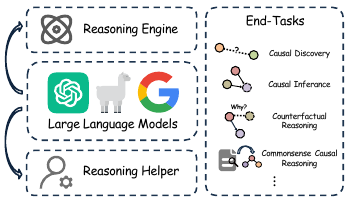}
  \caption{Large language models for causal reasoning: serving as reasoning engines or providing support to traditional methods in various end-tasks.}
  \label{fig:intro}
\end{figure}

\tikzstyle{my-box}= [
    rectangle,
    draw=hidden-draw,
    rounded corners,
    text opacity=1,
    minimum height=1.5em,
    minimum width=5em,
    inner sep=2pt,
    align=center,
    fill opacity=.5,
]
\tikzstyle{leaf}=[my-box, minimum height=1.5em,
    fill=pink!60, text=black, align=left,font=\scriptsize,
    inner xsep=2pt,
    inner ysep=4pt,
]
\begin{figure*}[t]

    \centering
    \resizebox{\textwidth}{!}{
        \begin{forest}
            forked edges,
            for tree={
                grow=east,
                reversed=true,
                anchor=base west,
                parent anchor=east,
                child anchor=west,
                base=left,
                font=\small,
                rectangle,
                draw=hidden-draw,
                rounded corners,
                align=left,
                minimum width=4em,
                edge+={darkgray, line width=1pt},
                s sep=3pt,
                inner xsep=2pt,
                inner ysep=3pt,
                ver/.style={rotate=90, child anchor=north, parent anchor=south, anchor=center},
            },
            where level=1{text width=3.5em,font=\scriptsize,}{},
            where level=2{text width=7.4em,font=\scriptsize,}{},
            where level=3{text width=5.6em,font=\scriptsize,}{},
            where level=4{text width=6.1em,font=\scriptsize,}{},
            [
                LLMs for Causal Reasoning, ver
                [
                Preliminaries\\ (\S \ref{Preliminaries})
                ]
                [
                    Methodology \\ (\S \ref{Towards Causal Reasoning in Large Language Models})
                    [
                        LLMs as  Reasoning \\Engines (\S \ref{Serve as Causal Reasoning Engines})
                            [
                            Fine-tuning \\ (\S \ref{Supervised Fine-tuning})
                            [                        
                            causalbert~\cite{li2021causalbert}{,} Causal Effect Tuning~\cite{zheng2023preserving}{,} 
                        SIT~\cite{chen2024unifiedcausalviewinstruction}{,}
                            \\LLM4Causal~\cite{jiang2023large}{,} meta-SCM ~\cite{anonymous2024a}
                                    , leaf, text width=25em
                            ]
                            ]
                            [
                            Prompt Engineering \\(\S \ref{Prompt Engineering})
                            [  
                            CausalCOT~\cite{jin2023cladder}{,} CausalCL~\cite{wang2023contextualized}{,} CREPE~\cite{zhang2023causal}{,}\\  
                            MCR~\cite{zang2023discovering}{,} CPL~\cite{he2022cpl}
                                    , leaf, text width=25em
                                ]
                                ]
                            [
                            Tools Integration \\(\S \ref{External Tool Integration})
                            [chatcot~\cite{chen2023chatcot}{,} Toolformer~\cite{schick2023toolformer}{,} PLAN ~\cite{lu2022neuro}{,} \\augmented-LLM ~\cite{pawlowski2023answering, komeili2021internetaugmented}
                                    , leaf, text width=25em
                                ]
                                ]
                            [
                            Alternative Approach \\(\S \ref{Others})
                            [Allure~\cite{hasanbeig2023allure}{,} LLM-debate~\cite{du2023improving}{,} NLE~\cite{atanasova2023faithfulness}{,} \\CausalGPT~\cite{tang2023causalgpt}{,} SI~\cite{creswell2022selectioninference, creswell2022faithful}{,} \\ LLM-cascade~\cite{yue2023large}
                                    , leaf, text width=25em
                                ]
                                ]
                        ]
                        [
                            LLMs as Helper to \\ traditional method (\S \ref{Prior})
                            [
                                Information \\Extractor (\S \ref{Information Extractor})
                                [
                                    CARING~\cite{yang2023neurosymbolic}{,}
                                    DML-LLM ~\cite{abdali2023extracting}{,}    RealTCD~\cite{li2024realtcdtemporalcausaldiscovery}
                                    , leaf, text width=25em
                                ]
                            ]
                            [
                                Data Generator \\(\S \ref{Data Generator})
                                [
                                    ILS-CSL~\cite{ban2023causal}{,}  CausalBank~\cite{ijcai2020-guided}{,} Adversarial~\cite{staliūnaitė2021improving}
                                    , leaf, text width=25em
                                ]
                            ]
                        ]
                    ]
                    [
                        Evaluation\\ (\S \ref{Measuring Casual Reasoning})
                        [
                            Tasks \& Benchmarks (\S \ref{Experiment})
                            [
                                 Causal Discovery~\cite{roemmele2011choice,mooij2015distinguishingcauseeffectusing,luo2016commonsense,khashabi2018looking,tu2019neuropathic}\\ \cite{ning2019joint,mostafazadeh2020glucosegeneralizedcontextualizedstory,du2022ecare,bondarenko2022causalqa,jin2023can}{,} \\Causal Inference~\cite{mostafazadeh-etal-2016-corpus,qin2019counterfactual,wang2023cola,jin2023cladder,romanou2023crab}{,} \\Additional Causal Tasks  ~\cite{sap2019socialiqa,frohberg-binder-2022-crass,nie2023moca,wang2023tram}
                                , leaf, text width=32.7em
                            ]
                        ]
                        [
                            Results (\S \ref{main results}) and Analysis\\ (\S \ref{analysis})
                        ]
                    ]
                ]
            ]
        \end{forest}
    }
    \caption{Structure overview. We categorize the role of LLMs in causal reasoning into two main functions: as reasoning engine and as helper. Each function is further divided into specific methodologies. We also outline the evaluation process, including tasks, benchmarks, results and analysis.}
    \label{categorization_of_reasoning}
\end{figure*}

Enhancing the causal reasoning capabilities of LLMs differs from improving their general reasoning skills. CR focuses on understanding causal relationships, while general reasoning primarily relies on identifying correlations \cite{bao2024llmschainofthoughtnoncausalreasoners}. Unlike general reasoning, which often involves simpler, more static relationships, CR requires deeper analysis and managing uncertainty to determine how events are causally linked \cite{pearl2009causality}. Additionally, CR must account for confounding variables \cite{10.1214/ss/1009211805}, where an observed correlation between two variables is due to the influence of a third variable. Counterfactual reasoning \cite{pearl2009causality}, often necessary for causality verification, is also challenging, as it involves reasoning about hypothetical scenarios that did not actually occur.

Existing research on enhancing the CR capabilities of LLMs can be broadly categorized based on their role: either as the \textbf{engine} directly performing causal inferences or as the \textbf{helper} providing extra information to traditional CR methods (Figure \ref{fig:intro}). Despite significant progress, challenges such as the shallow depth of causal reasoning skills and the scarcity of high-quality data still require further exploration \cite{zevcevic2023causal}. Addressing these challenges is crucial, as enhancing causal reasoning not only improves the general performance of LLMs but also provides substantial benefits for tackling complex reasoning problems and supporting decision-making \cite{liu2024largelanguagemodelscausal}. Thus, integrating causal inference into the model architecture \cite{melnychuk2022causal} or during the pre-training phase \cite{NEURIPS2023_642a321f} presents a promising direction for future research.

In this work, we focus on methodologies to enhance the causal reasoning capabilities of LLMs, exploring practical challenges and identifying potential opportunities. 
We introduce a fine-grained \textbf{taxonomy} (\S \ref{Towards Causal Reasoning in Large Language Models}) that systematically reviews and summarizes current techniques employed in LLMs to improve CR.
To evaluate these capabilities, we conduct extensive \textbf{assessments} (\S \ref{Measuring Casual Reasoning}) of widely-used LLMs on existing benchmarks, providing key findings and detailed analysis.
Additionally, we discuss the \textbf{limitations} of current approaches (\S \ref{Findings and Reflections}) and propose promising directions for \textbf{future research} (\S \ref{Future Directions}).
Previous work \citep{liu2024large, wan2024bridging} reviews causal reasoning in LLMs, focusing more on the perspective of \emph{traditional causal reasoning}.
In contrast, our focus lies in the \emph{methodological advancements} that enhance the causal reasoning abilities of LLMs.
More importantly, we conduct \emph{practical evaluations} of LLMs on various CR tasks, offering valuable experimental results and insights that are often lacking in related literature.
By bridging theoretical advancements and practical evaluations, our study fills a critical gap in the literature and offers insights to guide the development of LLMs with enhanced causal reasoning.

\section{Preliminaries}\label{Preliminaries}

CR in the context of machine learning (ML) involves understanding and modeling cause-and-effect relationships within data, which goes beyond the traditional ML focus on correlations. Traditional ML methods often identify patterns or correlations, limiting their applicability in understanding the mechanisms behind complex systems \cite{peters2017elements}. Causal ML, on the other hand, addresses this gap by focusing on cause-and-effect relationships, which is crucial for making informed decisions, predicting the effects of actions, and understanding complex systems \cite{prosperi2020causal, fernandez2022causal, lagemann2023deep, berrevoets2023causal}.

\begin{table}[t]\small\centering
    \newcolumntype{Y}{>{\centering\arraybackslash}X}
    \centering
    \begin{tabularx}{\columnwidth}{YYY}
        \toprule
        Levels          & Equation             & Query \\
        \midrule
        Association     & $P(y|x)$             & What is?          \\
        Intervention    & $P(y|\mathrm{do}(x),z)$ & What if?        \\
        Counterfactual  & $P(y_x|x',y')$       & Why?              \\
        \bottomrule
    \end{tabularx}
    \vspace{-2mm}
    \caption{The Ladder of Causality}
    \label{tab:causaltype}
\end{table}

\begin{figure*}[!t]
 \centering
  \includegraphics[width=1\textwidth]{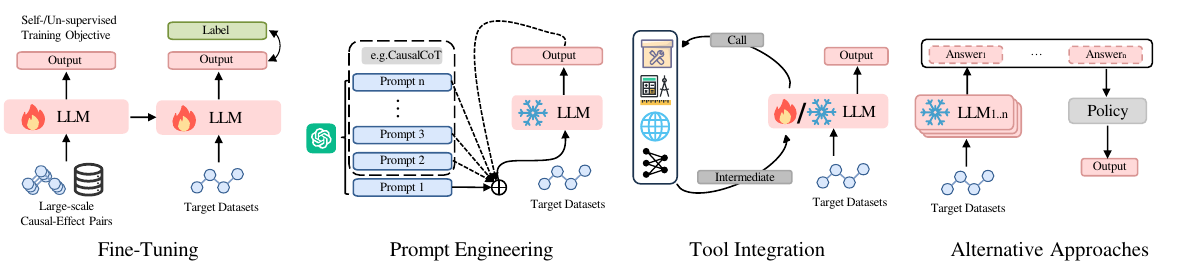}
  \caption{Overview of methods for LLMs as causal reasoning engines. (a) Fine-Tuning: Adapting LLMs using large-scale causal-effect pairs and target datasets.
(b) Prompt Engineering: Crafting targeted prompts to elicit the internal CR capabilities.
(c) Tool Integration: Leveraging external tools to support LLMs in performing intermediate steps.
(d) Alternative Approaches: Implementing additional methods, such as iterative improvement protocols, multi-agent systems, and rationale-based evaluation.}
  \label{fig:engine}
\end{figure*}

The CR framework can be categorized into three levels: association, intervention, and counterfactual \cite{pearl2000models, pearl2018book}, which form ``the ladder of causality'' (Table \ref{tab:causaltype}). Association is formalized in statistical terms through conditional probability between variables. Intervention is represented using Pearl's do-operator to explore causal effects, while counterfactual reasoning involves hypothetical scenarios that address potential outcomes under different circumstances. Structural Causal Models (SCMs), often visualized with Directed Acyclic Graphs (DAGs) where nodes represent variables and edges indicate causal relationships \cite{pearl2009causality}, are widely used to represent the interplay of these components. SCMs provide a mathematical framework for formalizing causal relationships within a system, thus serving as the foundation for causal reasoning and causal learning \cite{peters2017elements}. More details about CR tasks, and the comparison between LLMs and traditional methods, are provided in Appendix \ref{appendix:preliminaries}.

\section{Towards Causal Reasoning in Large Language Models}

\label{Towards Causal Reasoning in Large Language Models}

We separate the roles of language models in CR into two categories. First, LLMs can serve as \textbf{causal reasoning engines}, employing methods such as fine-tuning, prompt engineering, external tool integration, and alternative approaches to directly draw causal inferences from input prompts. 
Second, LLMs can act as \textbf{helpers} to traditional methods by extracting causal information and generating causality data to enhance causal analysis across various domains.

\subsection{Serving as Causal Reasoning Engines}
\label{Serve as Causal Reasoning Engines}

\subsubsection{Injecting Causal Knowledge through Fine-Tuning} \label{Supervised Fine-tuning} 

Fine-tuning language models has been proven to be a simple and effective technique for transferring learned knowledge from a pre-trained model to new downstream tasks. Compared to training from scratch, it can significantly reduce training time, and improve performance on new tasks ~\cite{lester2021power}. For example, \citet{li2021causalbert} used self-supervised learning strategies, such as causal pairs ranking and fine-tuning, to inject causal knowledge into pre-trained models. Additionally, \citet{zheng2023preserving} proposed Causal Effect Tuning, which is based on causal inference, to learn new knowledge from target data while preserving old knowledge from pre-trained models, thereby building a causal path without storing the pre-training data.

Although scaling language models increases the effectiveness of their prior knowledge in addressing complex causal questions, most training corpora collected from online sources consist of unstructured textual data rather than structured causal annotations.
To address this, \citet{jiang2023large} collected a custom dataset containing causal questions and interpreted intents in a structured format for supervised fine-tuning.
Building on this, \citet{chen2024unifiedcausalviewinstruction} combined the structural causal model with instruction tuning~\cite{liu2023llava,wei2022finetuned,chung2022scaling} to develop a meta Structural Causal Model, which unified different NLP tasks under a single causal structure and introduced the Structural Instruction Tuning method to learn task-specific causal representations that mimic the causal factors for each task.

\paragraph{Limitations}
Despite these progresses, LLMs often lack clearly defined supervision for causality, as most ML datasets contain only observational data without interventions, making key causal quantities, such as treatment effects, unknown ~\cite{kıcıman2023causal}. Furthermore, causality datasets often feature complex, heterogeneous relationships between variables ~\cite{huang2020causaldiscoveryheterogeneousnonstationarydata}, which are more challenging to capture compared to linguistic or perceptual patterns.
Thus, developing methods to bridge this gap, such as integrating causal structure learning or leveraging synthetic data for interventions, remains an important area of research.

\subsubsection{Embedding Causal Knowledge within Contexts}\label{Prompt Engineering} 

Prompt-based methods have become crucial in unlocking the internal reasoning capabilities of LLMs across various tasks. Compared with general reasoning, CR tasks, particularly multi-step ones, pose greater challenges due to factors such as the complexity of understanding cause-and-effect relationships, the need for reasoning about interventions and counterfactuals, and the presence of confounding variables \cite{gandhi2023understanding,tan2023evidence,ban2023causal,abdali2023extracting}. 

To address these challenges, various innovative prompting strategies have been proposed \cite{liu2023kept,bagheri2024c2pfeaturinglargelanguage}. Building on Chain-of-Thought (CoT) prompting \cite{wei2023chainofthoughtpromptingelicitsreasoning}, CausalCoT \cite{jin2023cladder} enables LLMs to reason step-by-step through causal relationships, significantly improving causal inference tasks involving multiple causes and effects. Recent advancements, such as Causal Contextualized Learning (CausalCL) \cite{wang2023contextualized}, and \citet{zhang2023causal} further enhances CR by combining code-like prompting with chain-of-thought prompting, boosting the performance for multihop event reasoning.
Multi-modal Causal Reasoning (MCR) \cite{zang2023discovering} extends CR to multi-modal settings, integrating visual and textual information for a comprehensive analysis. Counterfactual Prompting Learning (CPL) \cite{he2022cpl} utilizes counterfactual thinking to guide models in exploring alternative scenarios, thereby improving their understanding of causal mechanisms.

These advanced prompting techniques collectively enhance the ability of LLMs to understand causal relationships, thereby improving precision in CR tasks and enabling models to manage long-term dependencies more effectively \cite{lu2022neuro}. Furthermore, they reduce biases from training data, enrich the model's knowledge with external information and commonsense, and do so without substantial increases in computational costs \cite{yao2022react,jin2023cladder}.

\paragraph{Limitations} Causal prompt engineering depends on the model's inherent causality knowledge, which is often insufficient due to the correlation-driven nature of its modeling approach ~\cite{zhang2024probingcausalitymanipulationlarge}. Crafting effective prompts is labor-intensive, requiring significant domain expertise. Furthermore, the sensitivity to prompt variations can lead to inconsistent outputs, reducing robustness~\cite{perez2021truefewshotlearninglanguage}. Additionally, multi-step causal prompts can increase computational inefficiency, demanding more resources and time. These limitations highlight the necessity of complementary approaches, such as explicit causal modeling or specialized fine-tuning.

\subsubsection{External Causal Tool Integration}\label{External Tool Integration}

The motivation for integrating external tools into LLMs is that effective causal reasoning requires more than just internal model representations. By accessing structured knowledge bases and performing specialized computations, external tools enhance LLMs' capacity for complex causal reasoning, complementing their inherent capabilities. The ability to use these tools can be acquired through methods such as fine-tuning, in-context learning, and prompt manipulation. This integration establishes a crucial link between causal reasoning and language understanding, enabling LLMs to handle complex causal tasks more effectively.

For methods that incorporate tools in prompts, PLAN \cite{lu2023neurosymbolic} introduced a causal framework for procedural planning that leverages external knowledge bases, such as ConceptNet \cite{speer2018conceptnet}. It semantically parses tasks into an entity set to retrieve relevant sub-graphs for improved planning.
Additionally, \citet{pawlowski2023answering} compared context augmentation and tool augmentation methods: context augmentation uses language models for supplementary operations beyond the causal expert system, whereas tool augmentation incorporates basic Python tools to manipulate outputs from the expert system.
Their experiments showed that context-augmented LLMs were more prone to errors compared to tool-augmented ones. For methods involving supervised fine-tuning, dialogue agents such as those by \citet{shuster2022blenderbot, komeili2021internetaugmented} are trained to effectively use search engines.

\paragraph{Limitations} While integrating external causal tools can enhance LLMs, it also introduces several challenges. The effectiveness of these tools depends heavily on the quality and completeness of external resources, which can lead to errors when inadequate. The alignment between the model and tools is often complex and domain-specific, limiting the model's generalizability. Additionally, reliance on external tools may slow response times and increase the risk of errors, as observed with context-augmented models \cite{pawlowski2023answering}. These challenges indicate that external tools should be used in combination with other approaches to fully address complex CR tasks.

\subsubsection{Alternative Approaches}
\label{Others}

Recent studies have explored various alternative approaches to enhance the causal reasoning capabilities of LLMs, including iterative improvement protocols, multi-agent systems, and rationale-based evaluation. For example, ALLURE \cite{hasanbeig2023allure} uses a systematic auditing and \emph{iterative improvement protocol} with in-context learning examples to refine LLM evaluation. ECHO \cite{xie2023echo} introduces a visio-linguistic dataset based on real-world scenarios to evaluate event causality through human-centric reasoning.

\emph{Agentic frameworks} have also been employed for causal reasoning in LLMs. 
CausalGPT \cite{tang2023causalgpt} employs a multi-agent system comprising reasoners and evaluators, with reasoners generating answers and evaluators assessing factual accuracy and causal consistency.
\citet{du2023improving} presents a debate framework where multiple instances of LLMs debate a causality-related query to reach a consensus.
\emph{Selection and inference-based techniques} are another approach. \citet{creswell2022selectioninference} alternate between selection and inference steps using pre-trained LLMs to generate interpretable causal reasoning steps, while \citet{creswell2022faithful} chain reasoning steps to produce valid reasoning traces.
\emph{Rationale-based evaluation methods} are also receiving increased attention. \citet{atanasova2023faithfulness} use rationale-based methods to conduct tests for evaluating the faithfulness of natural language explanations. \citet{abdali2023extracting} generate a prior model to address domain knowledge gaps, serving as a heuristic for assessing the informativeness of user feedback. Additionally, \citet{yue2023large} utilize a cascaded model incorporating CoT and Program-of-Thought (PoT) approaches, along with answer sampling and consistency checking, to enhance causal reasoning.

\subsection{Enhancing Traditional CR Methods}
\label{Prior}

\begin{figure}[t]
 \centering
  \includegraphics[width=0.48\textwidth]{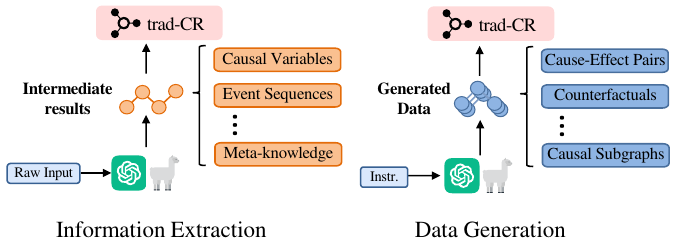}
\vspace{-2mm}
  \caption{Overview of methods for using LLMs to enhance traditional approaches. (a) Information Extraction: Extracting causal variables and events from text and adjusting for biases. (b) Data Generation: Generating synthetic causal data and forming hypotheses.}
  \label{fig:helper}
\end{figure}

Beyond serving as causal reasoning engines, LLMs also act as helpers in causal analysis across domains (Figure \ref{fig:helper}), fulfilling several supportive roles:
(1) \emph{Causal Information Extraction}: LLMs can extract causal variables, event sequences, and meta-knowledge from unstructured text, enhancing the quality of causal discovery in real-world applications ~\cite{yang2023neurosymbolic,abdali2023extracting,li2024realtcdtemporalcausaldiscovery}. Additionally, by leveraging prior knowledge, they identify and adjust for biases and confounders, improving the robustness of causal inferences~\cite{ban2023causal}.
(2) \emph{Causality Data Generation \& Augmentation}: LLMs generate synthetic, causally relevant data, such as cause-effect pairs, counterfactuals, and causal subgraphs, thereby enriching existing datasets and supporting traditional CR methods~\cite{staliūnaitė2021improving}. Furthermore, they form causal hypotheses from observed data patterns, generate plausible counterfactuals, and reason through complex interventions, guiding CR methods to explore less obvious causal pathways and enabling effective outcome evaluation ~\cite{gat2023faithful}.
These supportive roles illustrate the potential of LLMs to complement traditional causal analysis methods by bridging gaps in data, knowledge, and reasoning. Through their ability to extract hidden causal information and generate synthetic data, LLMs enable more robust and comprehensive causal reasoning.
More details on these supportive roles are provided in Appendix \ref{appendix:helper}.

\begin{table*}[h]\small
\centering 
\resizebox{\linewidth}{!}{%
\vspace{1cm}
\begin{NiceTabular}{l|cccc|cc|ccc}
    \toprule
    \multirow{3}{*}{\textbf{Model}} &\multicolumn{4}{c}{\textbf{Causal Discovery}}  &\multicolumn{2}{c}{\textbf{Causal Inference}}&\multicolumn{3}{c}{\textbf{Additional Causal Tasks}} \\ 
    \cline{2-10} & \textbf{COPA} & \textbf{NPDS} & \textbf{e-CARE} & \textbf{Corr2Cause}  & \textbf{CLADDER} & \textbf{CRAB} & \textbf{CRASS} & \textbf{MoCa} & \textbf{Tram} \\
    \midrule
    Mistral-7B & 86.2 \downnum{3.0} & 61.0 \downnum{10.0} & 63.4 \upnum{5.0} & 24.2 \upnum{4.6} & 55.0 \downnum{6.4} & 35.6 \downnum{1.4} & 56.7 \upnum{15.0} & 55.6 \upnum{3.4} & 75.8 \downnum{2.2}\\
 \qquad + CoT & 88.8 \downnum{7.4} & 65.0 \downnum{1.0} & 69.8 \upnum{0.8} & 36.6 \downnum{3.2} & 60.0 \downnum{2.4} & 39.0 \downnum{3.6} & 71.7 \downnum{1.7} & 53.5 \upnum{0.7} & 76.8 \downnum{3.4}\\

    Mistral-8$\times$7B & 75.2 \upnum{7.8} & 73.0 \downnum{22.0} & 58.6 \upnum{15.2} & {40.0} \downnum{9.0} & 57.8 \downnum{2.2} & 34.6 \downnum{8.8} & 58.3 \upnum{20.0} & 57.6 \downnum{2.7} & 67.6 \upnum{7.4}\\
 \qquad + CoT &82.8 \upnum{6.4} &65.0 \downnum{3.0} &72.0 \upnum{5.6} &33.6 \upnum{10.0} &58.6 \upnum{4.4} &35.6 \downnum{1.4} &71.7 \upnum{8.3} &{59.0} \upnum{2.1} &69.6 \upnum{6.0}\\

Gemma-2-9B & 91.2 \downnum{8.4} & 67.0 \downnum{4.0} & 69.4 \upnum{5.4} & 31.0 \downnum{1.2} & 56.8 \downnum{1.6} & 45.2 \downnum{2.6} & 76.7 \downnum{1.7} & 57.6 \upnum{2.1} & 79.2 \downnum{0.8}\\
\qquad + CoT & 96.0 \upnum{0.0} & 77.0 \downnum{26.0} & 80.6 \upnum{1.4} & 37.6 \upnum{2.4} & 56.0 \downnum{1.0} & 40.2 \downnum{2.8} & 76.7 \upnum{8.3} & 63.2 \downnum{3.5} & 83.2 \downnum{1.8}\\

Gemma-2-27B & 94.0 \downnum{6.5} & 74.0 \downnum{10.0} & 70.0 \upnum{8.0} & 23.8 \upnum{12.4} & 57.6 \upnum{6.4} & 47.4 \downnum{3.8} & 86.7 \downnum{6.7} & 54.9 \upnum{7.6} & 78.6 \upnum{1.6}\\
\qquad + CoT & 97.8 \downnum{0.6} & \textbf{80.0} \downnum{17.0} & 83.8 \downnum{1.0} & \textbf{42.4} \downnum{1.4} & 57.2 \upnum{6.2} & 45.4 \downnum{5.6} & 91.7 \downnum{5.0} & 58.3 \upnum{9.8} & 84.8 \downnum{3.2}\\

    LLaMA3-8B &81.4 \upnum{9.4} &57.0 \upnum{6.0} &62.8 \upnum{8.2} &36.0 \downnum{4.0} &53.0 \upnum{1.2} &31.4 \upnum{10.0} &61.7 \upnum{15.0} &58.3 \downnum{1.4} &68.6 \upnum{9.0}\\
 \qquad + CoT &89.8 \upnum{2.4} &56.0 \upnum{1.0} &76.0 \downnum{1.6} &27.2 \upnum{6.0} &56.8 \upnum{1.4} &35.6 \upnum{0.8} &76.7 \upnum{1.6} &51.4 \downnum{2.8} &78.2 \upnum{1.2}\\

    LLaMA3-70B &98.0 \upnum{0.8} &62.0 \upnum{4.0} &76.6 \upnum{4.8} &27.2 \upnum{4.8} &59.2 \upnum{12.2} &33.6 \upnum{5.0} &88.3 \upnum{0.0} &54.9 \upnum{5.5} &81.2 \downnum{2.0}\\
 \qquad + CoT &98.2 \upnum{0.4} & {76.0} \downnum{12.0} & {82.8} \upnum{2.0} &34.8 \upnum{3.2} &55.0 \upnum{20.8} &43.0 \upnum{1.0} &88.3 \upnum{6.7} &{59.0} \upnum{2.1} &83.4 \downnum{2.4}\\

    LLaMA3.1-8B &82.4 \upnum{12.2} &48.0 \upnum{12.0} &61.2 \upnum{14.6} &37.2 \downnum{14.6} &53.6 \downnum{3.8} &34.6 \upnum{8.8} &66.7 \upnum{16.6} &53.5 \downnum{9.1} &68.6 \upnum{9.2}\\
 \qquad + CoT &88.2 \upnum{6.8} &63.0 \downnum{3.0} &77.8 \upnum{1.4} &31.2 \downnum{4.6} &48.4 \upnum{4.4} &35.4 \downnum{4.0} &70.0 \upnum{5.0} &53.5 \downnum{0.7} &79.6 \downnum{2.2}\\

    LLaMA3.1-70B &98.2 \upnum{0.8} &63.0 \upnum{3.0} &78.2 \upnum{1.4} &23.2 \upnum{3.2} &59.2 \upnum{15.2} &37.2 \upnum{3.4} &88.3 \upnum{6.7} &56.9 \upnum{1.4} &80.4 \downnum{1.0}\\
 \qquad + CoT & 98.8 \upnum{0.8} &68.0 \downnum{5.0} &81.4 \upnum{4.4} &23.6 \upnum{13.2} &60.6 \upnum{10.2} &{44.6} \downnum{3.6} &\textbf{96.7} \downnum{3.4} &55.6 \upnum{2.7} &{83.8} \downnum{3.0}\\
DeepSeek-V3$^{*}$ & 98.0 \upnum{1.0} & 62.0 \upnum{7.0} & 83.0 \upnum{3.0} & 35.0 \upnum{6.0} & 62.0 \upnum{10.0} & 50.0 \upnum{5.0} & 95.0 \upnum{1.7} & 63.0 \upnum{2.0} & \textbf{87.0} \upnum{1.0} \\
\qquad + CoT$^{*}$ & \textbf{100.0} \upnum{0.0} & 64.0 \upnum{7.0} & \textbf{84.0} \upnum{8.0} & 37.0 \upnum{9.0} & 76.0 \upnum{8.0} & \textbf{56.0} \downnum{6.0} & \textbf{96.7} \upnum{0.0} & \textbf{69.0} \downnum{1.0} & 86.0 \upnum{2.0} \\
DeepSeek-R1$^{*}$ & \textbf{100.0} & 61.0 & \textbf{84.0} & 38.0 & \textbf{92.0} & 51.0 & 95.0 & 61.0 & 82.0 \\
    \midrule
    GPT-3.5-turbo &89.8 \upnum{3.4} &54.0 \downnum{5.0} &72.0 \downnum{2.2} &41.8 \upnum{8.4} &52.4 \upnum{6.8} &32.0 \upnum{6.0} &75.0 \upnum{10.0} &56.2 \downnum{1.3} &76.6 \downnum{0.6}\\
 \qquad + CoT &95.2 \upnum{1.2} &70.0 \downnum{9.0} &82.8 \downnum{4.2} &34.6 \downnum{0.6} &54.8 \upnum{4.4} &32.6 \upnum{11.6} &85.0 \upnum{5.0} &56.9 \upnum{2.1} &82.4 \downnum{1.0}\\

    GPT-4o$^{*}$ & \textbf{100.0} \upnum{0.0} &56.0 \upnum{11.0} &85.0 \upnum{1.0} &47.0 \upnum{2.0} &61.0 \upnum{18.0} &33.0 \upnum{4.0} &95.0 \upnum{3.3} &61.0 \downnum{4.0} &84.0 \upnum{1.0}\\
 \qquad + CoT$^{*}$ &99.0 \upnum{1.0} &\textbf{73.0} \downnum{1.0} &\textbf{90.0} \downnum{3.0} &49.0 \upnum{4.0} &69.0 \upnum{10.0} &42.0 \upnum{3.0} &\textbf{100.0} \downnum{1.7} &61.0 \downnum{1.0} &87.0 \upnum{1.0}\\
    Claude 3.5$^{*}$ &98.0 \upnum{1.0} &63.0 \upnum{5.0} &72.0 \upnum{12.0} &45.0 \upnum{2.0} &70.0 \upnum{13.0} &57.0 \downnum{1.0} &95.0 \upnum{3.3} &42.0 \upnum{20.0} &84.0 \upnum{0.0}\\
 \qquad + CoT$^{*}$ &99.0 \upnum{1.0} &71.0 \upnum{0.0} &86.0 \upnum{4.0} &35.0 \upnum{8.0} &72.0 \upnum{9.0} &\textbf{58.0} \upnum{2.0} &98.3 \downnum{3.3} &59.0 \upnum{5.0} &\textbf{88.0} \downnum{3.0}\\
o1$^{*}$ &100.0  &56.0  &80.0  &59.0  &89.0  &40.0  &96.7  &60.0  &84.0 \\ 
o3-mini &99.8  &56.0  &79.6  &\textbf{60.6}  &\textbf{92.2}  &32.2  &93.3  &\textbf{63.9}  &81.4 \\
    \midrule
    Human  & 95.8 & 97.7 & 92.0 & 94.5 & 94.8 & 98.7 & 98.2 & 92.0 & 98.8\\
  
    \bottomrule
\end{NiceTabular}}
\vspace{-2mm}
\caption{Overall performance comparison of models across causal tasks. Performance is measured using pass@1 accuracy, with the best results in each category in bold. Zero-shot scores are used as the baseline, while the green and blue arrows indicate the change brought by few-shot learning. {$^*$}Model is evaluated on 100 random samples due to cost and time constraints.}
\label{tab:main-results}
\end{table*}

\section{Evaluating Casual Reasoning in Large Language Models}

\label{Measuring Casual Reasoning}

\subsection{Experimental Setup} 
\label{Experiment}
We first categorize the end tasks into three groups: causal discovery, causal inference, and additional causal tasks. 
For each category, we evaluate recent LLMs with strategies such as zero-shot, few-shot, direct I/O prompting, and Chain-of-Thought (CoT) reasoning.
To reduce variance, we evaluate each question 5 times using greedy decoding and compute the average accuracy (pass@1) for that question. We then calculate the overall average accuracy across all problems.
Specifically, we use COPA ~\cite{roemmele2011choice}, NPDS ~\cite{tu2019neuropathic}, e-CARE ~\cite{du2022ecare}, and Corr2Cause ~\cite{jin2023can} for causal discovery; CLADDER ~\cite{jin2023cladder} and CRAB ~\cite{romanou2023crab} for causal inference; and CRASS ~\cite{frohberg-binder-2022-crass}, MoCa ~\cite{nie2023moca}, and Tram ~\cite{wang2023tram} for additional causal tasks. 
Further details on task definitions, dataset statistics and examples, and implementation details are provided in Appendix \ref{appendix:dataset_overview} and Appendix \ref{appendix:implementation_details}.

\begin{figure}[t]
    \centering
    \includegraphics[width=\linewidth]{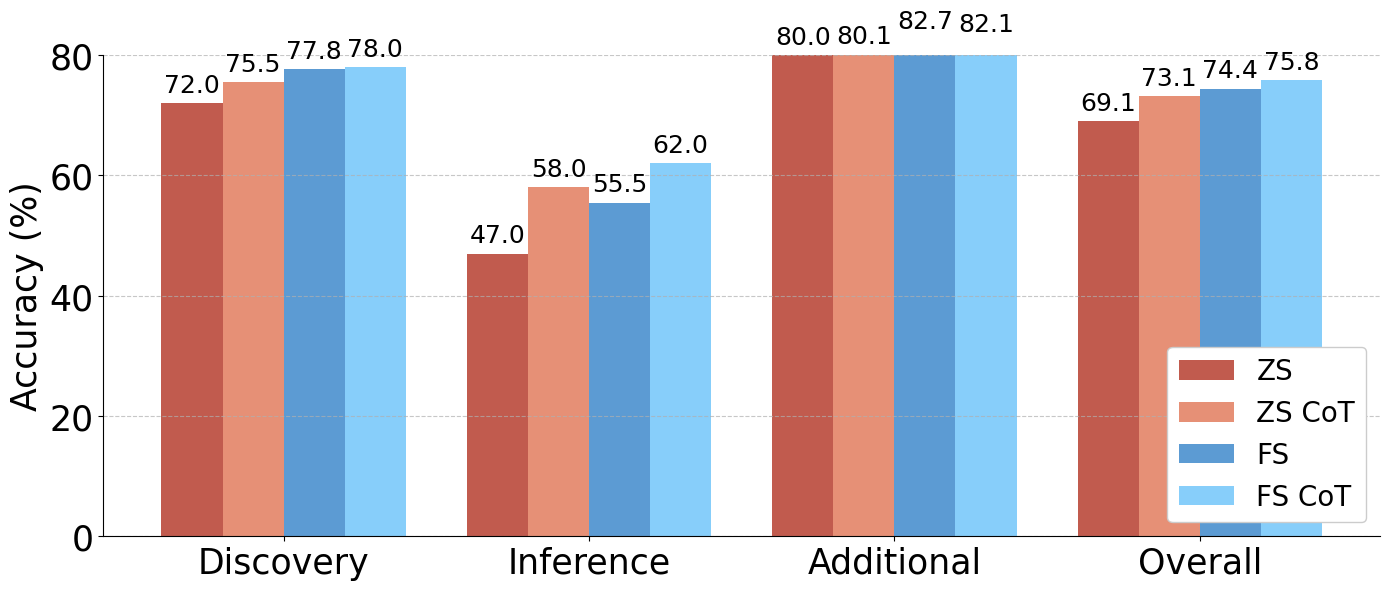}
    \caption{
    Performance gap of zero-shot (ZS) and few-shot (FS) learning, with and without CoT prompting. The results are from the best performing model (GPT-4o).}
    \label{fig:performance_comp}
\end{figure}

\subsection{Main Results}
\label{main results}

\begin{figure}[t]
    \centering
    \includegraphics[width=\linewidth]{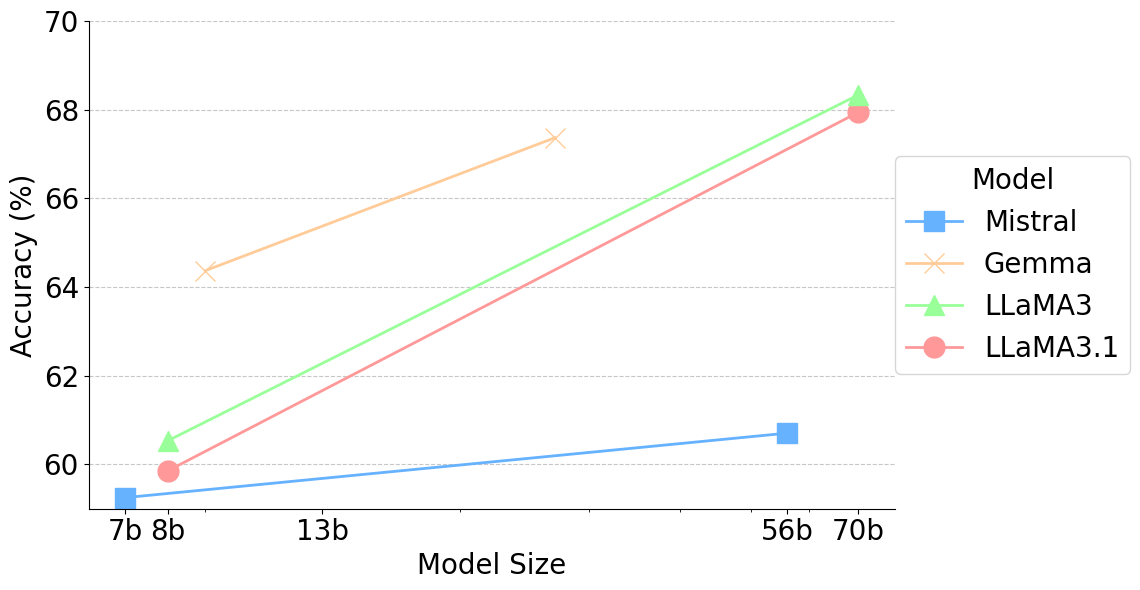}
\vspace{-5mm}
    \caption{
    Scaling effect of model size and overall casual reasoning performance. The x-axis (model size) is shown in the log scale. Mistral 8 × 7b (MoE) is treated as equivalent to 56b.}
    \label{fig:scaling_effect}
\end{figure}

Overall performance is shown in Table \ref{tab:main-results}. We summarize the key observations as follows:

\begin{table*}[h]\small
\centering 
\resizebox{\linewidth}{!}{%
\begin{NiceTabular}{l|cccc|cc|ccc}
    \toprule
    \multirow{3}{*}{\textbf{Method}} 
      & \multicolumn{4}{c}{\textbf{Causal Discovery}}  
      & \multicolumn{2}{c}{\textbf{Causal Inference}}
      & \multicolumn{3}{c}{\textbf{Additional Causal Tasks}} \\ 
    \cline{2-10}
      & \textbf{COPA} 
      & \textbf{NPDS} 
      & \textbf{e-CARE} 
      & \textbf{Corr2Cause}  
      & \textbf{CLADDER} 
      & \textbf{CRAB} 
      & \textbf{CRASS} 
      & \textbf{MoCa} 
      & \textbf{Tram} \\
    \midrule
    GPT-3.5-turbo 
      & 89.8  & 54.0  & 72.0  & 41.8  
      & 52.4  & 32.0  & 75.0  & 56.2  & 76.6 \\
    \quad + CoT 
      & 95.2  & 70.0  & 82.8  & 34.6  
      & 54.8  & 32.6  & 85.0  & 56.9  & 82.4 \\
    \quad + causalCoT$^{*}$
      & 49.0  & 28.0  & 37.0  & 35.0  
      & 45.0  & 19.0  & 52.0  & 45.0  & 48.0 \\
    \quad + FunctionCall$^{*}$ & 78.0 & 58.0 & 63.0 & 35.0 & 57.0 & \textbf{44.0} & 68.3 & 53.0 & 75.0 \\
    \midrule
    GPT-4o$^{*}$
      & \textbf{100.0} & 56.0  & 85.0  & 47.0  
      & 61.0           & 33.0  & 95.0  & \textbf{61.0} & 84.0 \\
    \quad + CoT$^{*}$ 
      & 99.0  & \textbf{73.0} & \textbf{90.0} & \textbf{49.0} 
      & 69.0  & 42.0 & \textbf{100.0} & \textbf{61.0} & \textbf{87.0} \\
    \quad + causalCoT$^{*}$
      & 94.0  & 58.0  & 75.0  & 47.0  
      & 75.0  & 36.0  & 93.0  & 56.0  & 78.0 \\
    \quad + FunctionCall$^{*}$
      & \textbf{100.0} & \textbf{73.0} & 73.0  & \textbf{49.0}  
      & \textbf{76.0} & 42.0  & 98.3  & \textbf{61.0}  & 84.0 \\
    \midrule
    Human  
      & 95.8  & 97.7  & 92.0  & 94.5  
      & 94.8  & 98.7  & 98.2  & 92.0  & 98.8\\
    \bottomrule
\end{NiceTabular}
}
\caption{Performance comparison between Direct IO, CoT, casualCOT prompting and functionCall (Python script) on GPT-3.5-turbo and GPT-4o across various causal reasoning datasets. Performance is measured using pass@1 accuracy, with the best results in each category in bold. {$^*$}Model is evaluated on 100 random samples due to cost and time constraints.}
\label{tab:comparison-improving-methods}
\end{table*}

\begin{figure*}[!t]
    \centering
    \includegraphics[width=0.98
\linewidth]{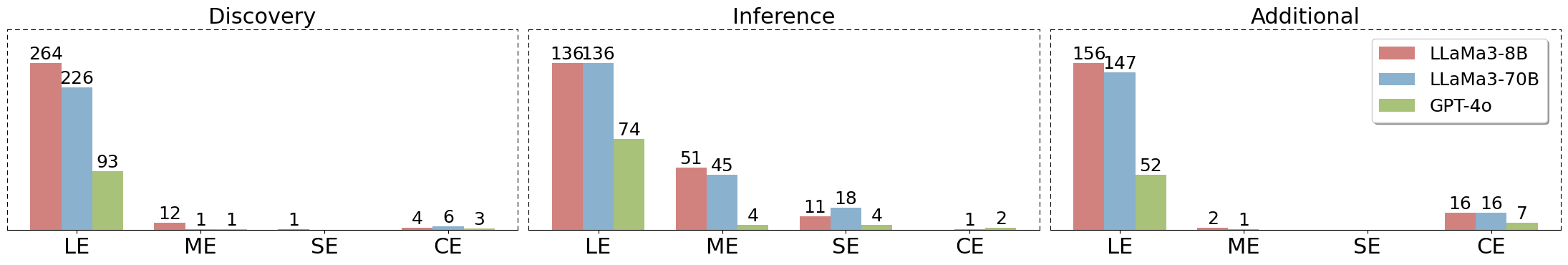}
    \caption{Error analysis across different types of causal reasoning tasks. For each task type, multiple datasets are analyzed with 100 randomly selected incorrect test samples. The y-axis shows the number of incorrect samples per error type: LE (Logical Errors), PE (Probabilistic Errors), SE (Statistical Errors), and CE (Contextual Errors). Detailed definition of each error type is given in Appendix \ref{appendix:implementation_details}.}
    \label{fig:error_analysis}
\end{figure*}

\noindent \textbf{LLMs significantly underperform humans in many CR Tasks.}
Our results reveal a notable performance gap between LLMs and humans across many causal reasoning tasks, as illustrated in the Radar chart (Figure \ref{fig:radarchart}, Appendix \ref{appendix:implementation_details}). Despite recent advancements, LLMs continue to face challenges in fully grasping and applying causal reasoning, often relying on surface-level patterns and struggling with multi-step reasoning. These limitations reduce their effectiveness in complex scenarios. Future work should focus on enhancing LLMs' ability to handle intricate causal structures, potentially through hybrid methods that integrate symbolic reasoning with statistical approaches. Addressing these challenges will be critical for bridging the gap between LLMs and human performance in causal reasoning.

\noindent \textbf{Performance disparity exists between discovery and inference tasks due to contextual and multi-variable reasoning demands.} Our results reveal distinct performance trends across causal tasks. All models achieve high accuracy in certain discovery tasks, such as COPA and e-CARE, which require  identifying potential causal relationships from observational data. However, their performance drops significantly in inference tasks, particularly when assessing causal strengths from long contexts (CRAB). Discovery tasks involving complex multi-variable interactions, such as Corr2Cause, also present significant challenges. This disparity underscores a fundamental limitation: while models are proficient at identifying potential causal relation candidates, they struggle with evaluating causal strengths and navigating intricate variable dependencies.

{
\noindent \textbf{Deep thinking models exhibit targeted gains in structured causal reasoning.}
Our analysis reveals that deep thinking models (DeepSeek-R1, o1, o3-mini) demonstrate particularly strong performance in structured causal reasoning tasks, especially in CLADDER where they achieve remarkable accuracy (92.0\% and 92.2\% respectively). This suggests that extended thinking contexts may be particularly beneficial for formal causal inference and graph-based reasoning. However, this advantage does not consistently extend to other causal tasks - performance on benchmarks such as COPA, NPDS, and e-CARE remains comparable to or worse than that of standard models. This trend indicates that deep thinking approaches primarily enhance structured causal reasoning rather than general causal understanding.
}

\subsection{Analysis and Discussion}
We further analyze the impact of various factors on accuracy below.
\label{analysis}

\noindent \textbf{Zero-shot or few-shot under CoT prompting.}
Figure \ref{fig:performance_comp} shows that both zero-shot and few-shot models benefit from CoT prompting, with few-shot CoT achieving the highest overall performance. While few-shot consistently outperforms zero-shot across all tasks, the models struggle the most with inference tasks, where accuracy remains lower than in discovery and additional tasks. These findings highlight the effectiveness of combining example-based learning with structured reasoning but also underscore the need for further improvements, especially in inference tasks, to enhance model performance across complex causal scenarios.

\noindent \textbf{Model size vs. Causal reasoning performance.}  
Figure \ref{fig:scaling_effect} demonstrates that causal reasoning performance improves consistently with increasing model size across all evaluated models. Both LLaMA3 and LLaMA3.1 achieve significant gains, with Gemma also exhibiting strong improvements. In contrast, Mistral shows only modest improvements as model size increases, likely due to its Mixture of Experts (MoE) architecture. These results underscore the importance of both model size and architecture in advancing causal reasoning capabilities.

\noindent \textbf{Error analysis.}
Through systematic analysis of model outputs, we identify four common types of errors in causal reasoning. 

\emph{Logical errors} occur when the model makes flawed cause-effect conclusions, such as confusing correlation with causation or failing to distinguish between necessary and sufficient conditions. For example, when asked whether someone caused a car to start in a scenario with multiple people each bringing a working battery, the model incorrectly attributes causation based solely on temporal sequence rather than considering the true causal relationship.

\emph{Probabilistic errors} arise when the model incorrectly handles numerical reasoning about likelihood and chance, including mishandling of conditional probabilities and confounding variables. In probability comparison tasks, the model often makes computational errors or misinterprets statistical dependencies, leading to incorrect conclusions about relative likelihoods.

\emph{Statistical errors} manifest when the model fails to properly account for complex relationships between multiple factors, such as indirect effects and selection bias. For instance, in scenarios involving multiple causal paths, the model often overlooks intermediate factors or assumes direct relationships where more complex dependencies exist.

\emph{Contextual errors} occur when the model fails to incorporate relevant background information or subtle details that should affect its causal reasoning. This often results in oversimplified conclusions that ignore important contextual factors or conditions that would alter the causal relationship.

Our analysis (Figure \ref{fig:error_analysis}) shows that logical errors are the most prevalent type of failure. Potential approaches to address these challenges include enhanced representation of causal relationships in the model architecture, systematic consistency checks for reasoning processes, and diversified training data covering a broader range of causal scenarios. Detailed examples of each error type and analysis of these approaches are provided in Appendix \ref{appendix:implementation_details}.

{
\noindent \textbf{Comparison with enhancement methods.}
We evaluated two methods for improving causal reasoning: CausalCoT \cite{jin2023cladder} and Function Call. CausalCoT provides structured steps for causal inference with domain-specific guidance, while Function Call leverages Python scripts for precise mathematical computations. Our results show that effectiveness varies by task type. For formal reasoning tasks involving causal graph analysis, CausalCoT outperforms other approaches due to its structured inference steps (75.0\% on CLADDER). In contrast, Function Call excels in numerical tasks, particularly on CLADDER (GPT-4o: 76.0\% vs 69.0\% with vanilla CoT), and performs comparably well on other computation-heavy datasets. However, for commonsense causal reasoning tasks like COPA and e-CARE, both methods underperform: CausalCoT shows substantial drops (40.8\% and 35\% respectively for GPT-3.5-turbo), while Function Call provides minimal gains (COPA: 78.0\% vs 95.2\% with vanilla CoT), suggesting that vanilla CoT remains the most effective for these tasks. These findings highlight the importance of task-aware method selection: Function Call for quantitative analysis, CausalCoT for structured causal inference with graphs, and vanilla CoT for commonsense causal reasoning.
}

\section{Findings and Reflections}
\label{Findings and Reflections}

\noindent \textbf{LLMs demonstrate shallow causal reasoning skills.}
Evaluation metrics often emphasize task-specific accuracy, leaving open questions about the depth of the model's causal reasoning \cite{zevcevic2023causal}.
For instance, \citet{hong2023large} investigated how LLMs process event {B} in a script-based story, which causally depends on a prior event {A}. While large models performed better than smaller ones in the $\neg \mathrm{A} \rightarrow \mathrm{B}$ condition, all models (even fine-tuned with code and instructions) failed to replicate human behavior when the cause was omitted ($\text{nil}$), indicating that LLMs lack sufficient representation of commonsense CR.

\noindent \textbf{Integration/Collaboration enhances the performance.}
Tasks that challenge a single LLM can often be addressed by integration with traditional methods, external tools, or multiple LLMs ~\cite{tang2023causalgpt,du2023improving}. 
Recent approaches, such as neuro-symbolic methods \cite{yang2023neurosymbolic} and iterative processes \cite{ban2023causal}, integrate LLMs with statistical analysis and causal structure learning, improving the quality of causal inferences.
The integration with causal analysis tools increases LLMs' access to causal knowledge across domains ~\cite{kıcıman2023causal}. As intermediaries between specialized tools and natural language, LLMs make causal reasoning more intuitive and user-friendly. 
This collaboration fosters tools for comprehensively exploring causal relationships in complex systems.

\noindent \textbf{Diverse and standardized causal reasoning benchmarks are needed.}
Comprehensive evaluation of the causal reasoning abilities of LLMs requires more diverse and standardized benchmarks. Existing benchmarks tend to be relatively simple and domain-specific. Future benchmarks should encompass a broader spectrum of causal reasoning challenges, including multi-step tasks with increasing complexity, while maintaining consistent causal structures. Such datasets would enable LLMs to explore a wider range of causal reasoning scenarios and better adapt to real-world applications ~\cite{yang2024criticalreviewcausalreasoning}.

\section{Future Directions}
\label{Future Directions}

\noindent \textbf{Deep understanding of existing theories.}  
Integrating causal reasoning capabilities into LLMs presents unique challenges, especially when requiring adherence to established theories. These theories, such as the potential outcomes framework, graph-based causal inference methods, and structural equation modeling, often rely on foundational assumptions, including the stable unit treatment value assumption, ignorability/unconfoundedness, and positivity ~\cite{liu2024large}. Considering these assumptions necessitates innovative strategies, such as incorporating specific post-training data and implementing filtering mechanisms to ensure the model's alignment with the theories.

\noindent \textbf{Data efficiency for end-to-end learning.}
LLMs have the potential to perform end-to-end causal reasoning, but their effectiveness depends on substantial data to learn and infer causal relationships. A major challenge is the scarcity of high-quality causality data, particularly for counterfactual reasoning, where alternative scenarios are often unavailable or difficult to generate. Improving data efficiency is therefore a critical research direction. Techniques such as analogical reasoning, imaginative simulation, and inferential methods can maximize the utility of existing data by generating synthetic datasets, filling data gaps, or extrapolating from limited information. These approaches enhance the model's capacity to learn causal relationships with constrained training data.

\noindent \textbf{Causality-based internal mechanisms for LLMs.}
Incorporating causal reasoning into LLMs internally is supposed to enhance their accuracy and depth in complex problem-solving scenarios \citep{kıcıman2023causal}. Thus, it is crucial to find feasible implementation ways so that LLMs generate content grounded in causality. One strategy could be the introduction of a dual-network architecture, where one is similar to current model and the other focuses on learning causal relations. Both networks are jointly trained to perform various tasks. Another approach involves modifying attention mechanisms to prioritize causal relevance. 
By embedding internal causal mechanisms, we believe LLMs can enhance their reasoning, address fairness and safety, and better handle multi-modal inputs through improved understanding of causal relations, leading to more accurate and transparent decisions and reduced biases.

\section{Conclusion}
\label{Conclusion}

LLMs equipped with these advanced technologies demonstrate strong potential in causal reasoning, but further research is needed to fully realize and refine their capabilities. Addressing challenges such as managing intricate causal structures is essential for continued development. Enhancing interpretability and transparency through causal reasoning is crucial for building trust and helping users rely on model inferences. Exploring innovative methods will be key to overcoming current limitations and unlocking the full potential of LLMs in complex, real-world scenarios. Advancing these areas will make LLMs more reliable and effective tools for causal analysis across domains.

\section*{Limitations}
In this survey, we review advancements in causal reasoning with LLMs, though several limitations should be noted. Our focus is primarily on text-based causal reasoning, with limited exploration of multi-modal or visual tasks, which restricts the generalizability of our findings across modalities. Although we evaluate mainstream LLMs and methods on popular benchmarks, we do not cover all enhancement techniques, particularly those requiring post-training or tailored to specific tasks and domains. Additionally, understanding the fundamental principles of causal reasoning in LLMs is crucial for ensuring trust and accountability in real-world applications, yet this area remains challenging and underexplored. Further research is warranted to address these gaps.

\bibliography{custom}

\appendix

\clearpage

\section{Casual Reasoning Tasks}
\label{appendix:preliminaries}

Causal discovery and inference \cite{glymour2019review,pearl2000models,spirtes2000causation,spirtes2016causal,pearl2016causal,perry2022causal,yao2021survey} are two related fields within various domains such as statistics and machine learning that aim to understand the relationships between variables in a system and identify cause-and-effect connections. 
\emph{(1) Causal discovery }\cite{glymour2019review,zhu2019causal,heckerman2006bayesian,huang2018generalized} aims to identify causal relationships from data, especially when the causal structure is unclear. It involves constructing models, such as causal graphs, to elucidate cause-and-effect. This is crucial as real-world constraints often make controlled experiments to establish causality impractical, unethical, or impossible \cite{spirtes2000causation,vemuri2015causal,peters2017elements}.
\emph{(2) Causal inference} \cite{pearl2009causality,yao2021survey,shams2022bayesian,pearl2016causal,mooij2016distinguishing,kuipers2022interventional,peters2017elements} involves adding data or interventions to a causal graph from causal discovery to test variable cause-effect relationships. It aims to determine the direct impact of a variable or intervention on another. The methods used for causal discovery may include constraint-based algorithms \cite{tsagris2018constraint,cooper1997simple,chaudhary2018community}, score-based algorithms\cite{huang2018generalized,peters2014identifiability,chickering2002optimal,heckerman2006bayesian}, hybrid algorithms \cite{tsamardinos2006max} and so on \cite{zanga2022survey,glymour2019review}. Constraint-based algorithms reconstruct causal graphs using conditional independence statements from statistical tests. Score-based methods optimize the fitness of the graph using score metrics from observed data. Hybrid approaches integrate both to enhance causal inference.

Although there has been much research based on traditional methods, they have several limitations in data requirements, model scalability, and model robustness that LLMs can address more effectively. 
\emph{(1) Data Quality:}
\textbf{Traditional Methods} require well-developed, high-quality datasets, making them vulnerable to missing data and measurement errors. They rely on assumptions like linearity, stationarity, and causal sufficiency, which may not hold in real-world scenarios.
\textbf{LLMs} capture non-linear interactions and are suitable for various types of data. They can augment sparse or incomplete data through in-context learning, generating plausible hypotheses, and filling gaps.
\emph{(2) Scalability:}
\textbf{Traditional Methods} can be computationally intensive and less scalable with increasing variables, often requiring costly intervention data.
\textbf{LLMs} efficiently process large datasets and automate causal discovery, making them more scalable and adaptable to complex datasets.
\emph{(3) Robustness:}
\textbf{Traditional Methods} simulate counterfactuals and predict intervention effects without explicit data, but their reliance on strict assumptions can limit robustness in practical applications.
\textbf{LLMs} apply methods such as fine-tuning to dynamically integrate vast background knowledge, providing a richer context for causal reasoning and making them more robust to data quality issues.

\section{Enhancing Traditional CR Methods}
\label{appendix:helper}

\subsection{Causal Information Extraction}\label{Information Extractor}

LLMs can directly support causal reasoning by serving as knowledge bases or extracting causal information. The effectiveness of problem-solving algorithms can be improved by adopting a modular approach that separates the logic and control components~\cite{10.1145/359131.359136}. The logic component represents the essential knowledge for solving the problem, while the control component manages the problem-solving strategy. Decoupling these components enhances flexibility, allowing updates to the control strategy without altering the underlying knowledge base.

\citet{yang2023neurosymbolic} proposed a neuro-symbolic approach with two components: an LLM-based symbolic representation generator that translates natural language into formal knowledge for symbolic inference, and an LLM-free symbolic inference engine. Similarly, \citet{abdali2023extracting} leveraged LLMs' reasoning capabilities to extract causal variables and event sequences, providing a heuristic for assessing feedback informativeness in the absence of domain knowledge. Additionally, RealTCD~\cite{li2024realtcdtemporalcausaldiscovery} introduced LLM-guided meta-initialization to extract meta-knowledge from textual information, improving the discovery process in real industrial settings.

Another notable contribution is the iterative LLM-supervised causal structure learning (ILS-CSL) framework introduced by \citet{ban2023causal}, which iteratively refines causal Directed Acyclic Graphs through feedback from LLMs. This framework highlights the efficiency and robustness of LLMs in improving causal discovery. 
More broadly, the ability of LLMs to extract causal information can reduce the dependency on purely data-driven methods and unlock new opportunities for hybrid CR approaches. As LLMs evolve, they are likely to play an increasingly important role in causal discovery and inference by connecting unstructured text with structured causal models, making CR more robust and adaptable across domains.

\begin{table*}[t!]
\centering

\resizebox{0.85\linewidth}{!}{%
\begin{NiceTabular}{lcccccc}
\toprule
\textbf{Dataset}  &\textbf{Size (Unit)}&\textbf{Domain}&\textbf{Source}&\textbf{Association}&\textbf{Intervention} &\textbf{Counterfactual} \\
\hline
\specialrule{0em}{1pt}{1pt}
\multicolumn{7}{c}{Causality Discovery}
\\ 
\midrule
SemEval Tasks 8 (\citeyear{hendrickx-etal-2010-semeval}) & 1.3K (P) & Mixed & Real & \Checkmark  & &   \\
\textbf{COPA} (\citeyear{roemmele2011choice}) & 1K (Q) & Dailylife & Real & \Checkmark  & &   \\
CEPairs (\citeyear{mooij2015distinguishingcauseeffectusing}) & 108 (P) & Mixed & Real & \Checkmark  & &  \\
CausalNet (\citeyear{luo2016commonsense}) & 62M (P) & Dailylife & Synthetic & \Checkmark  & &   \\
MultiRC (\citeyear{khashabi2018looking}) & 9.8K (Q) & Mixed & Real & \Checkmark  & &   \\
\textbf{NPDS} (\citeyear{tu2019neuropathic}) & 770 (R) & Health & Synthetic & \Checkmark  & &   \\
TCR Benchmark (\citeyear{ning2019joint}) & 25 (D) & News & Real & \Checkmark  & &   \\
SemEval Tasks 5 (\citeyear{yang-2020-semeval-task5}) & 20K (S) & Mixed & Real & \Checkmark  & & \Checkmark   \\
CausalBank (\citeyear{ijcai2020-guided}) & 314M (P) & Mixed & Synthetic & \Checkmark  & &   \\
GLUCOSE (\citeyear{mostafazadeh2020glucosegeneralizedcontextualizedstory}) & 4.8K (S) & Mixed & Real & \Checkmark  & &   \\
CausaLM (\citeyear{feder2021causalm}) & 9.8K (S) & Ethnic \& Gender & Real & \Checkmark  & & \Checkmark   \\
\textbf{e-CARE} (\citeyear{du2022ecare}) & 21K (Q) & Mixed & Real & \Checkmark  & & \Checkmark   \\
CausalQA (\citeyear{bondarenko2022causalqa}) & 1.1M (Q) & Mixed & Real & \Checkmark  & &   \\
\textbf{Corr2Cause} (\citeyear{jin2023can}) & 200K (S) & Mixed & Synthetic & \Checkmark  & &   \\
\hline
\specialrule{0em}{1pt}{1pt}
\multicolumn{7}{c}{Causal Inference}
\\ 
\midrule
ROC Stories (\citeyear{mostafazadeh-etal-2016-corpus}) & 100K (S) & Mixed & Real & \Checkmark  & &   \\
CSR (\citeyear{qin2019counterfactual}) & 16K (S) & Mixed & Real & \Checkmark  & \Checkmark & \Checkmark   \\
ART (\citeyear{bhagavatula2019abductive}) & 20K (S) & Mixed & Real & \Checkmark  & &   \\
CausalQA (\citeyear{bondarenko2022causalqa}) & 1.1M (Q) & Mixed & Real & \Checkmark  & &   \\
COPES (\citeyear{wang2023cola}) & 1.3K (P) & Mixed & Real & \Checkmark  & \Checkmark &   \\
\textbf{CLADDER} (\citeyear{jin2023cladder}) & 10K (S) & Mixed & Synthetic & \Checkmark  & \Checkmark & \Checkmark   \\
\textbf{CRAB} (\citeyear{romanou2023crab}) & 2.7K (P) & Dailylife & Real & \Checkmark  & &   \\
\hline
\specialrule{0em}{1pt}{1pt}
\multicolumn{7}{c}{Additional Causal Tasks}
\\ 
\midrule
SWAG (\citeyear{zellers2018swag}) & 113K (Q) & Mixed & Real & \Checkmark  & &   \\
CommonsenseQA (\citeyear{talmor2018commonsenseqa}) & 12K (Q) & Mixed & Real & \Checkmark  & &   \\
ATOMIC (\citeyear{sap2019atomic}) & 24K (S) & Mixed & Real & \Checkmark  & &   \\
Cosmos QA (\citeyear{huang2019cosmos}) & 35K (Q) & Mixed & Real & \Checkmark  & & \Checkmark   \\
SocialIQA (\citeyear{sap2019socialiqa}) & 38K (Q) & Social & Real & \Checkmark  & &   \\
\textbf{CRASS} (\citeyear{frohberg-binder-2022-crass}) & 275 (Q) & Mixed & Real & \Checkmark  & & \Checkmark   \\
\textbf{MoCa} (\citeyear{nie2023moca}) & 144 (S) & Mixed & Real & \Checkmark  & &   \\
\textbf{Tram} (\citeyear{wang2023tram}) & 1.2K (Q) & Dailylife & Real & \Checkmark & &   \\
\bottomrule
\end{NiceTabular}}
\caption{Benchmarks for causal reasoning tasks. Datasets used in our experiments are in bold. Sizes are reported in various units: P (causal pairs), R (causal relations), S (samples), Q (questions), and D (documents). We also provide their inclusion of the three levels of causality: association, intervention, and counterfactual.}
\label{tab:dataset_statistics}
\end{table*}

\subsection{Causality Data Generation \& Augmentation}\label{Data Generator}

LLMs store a vast amount of knowledge within their weights, and several works have explored how to extract causal-related knowledge to support traditional causal reasoning models. For instance, \citet{ijcai2020-guided} introduced a method where, given an input cause sentence, a causal subgraph is retrieved from an event causality network, and the most probable effect event is selected to guide the generation of an effect sentence. 
As LLMs scale, their capacity to generate causal alternatives has improved. For example, \citet{staliūnaitė2021improving} used GPT-2 to generate alternatives and re-write filtered augmentation data for causal relations. Similarly, \citet{gat2023faithful} showcased how LLMs can generate counterfactual approximations efficiently, focusing on model-agnostic explanations that enhance the interpretability and faithfulness of CR methods.

However, despite the impressive knowledge stored within LLMs, they remain prone to hallucinations, often generating incorrect or misleading facts ~\cite{wang2023surveyfactualitylargelanguage}. This issue is particularly problematic in causal reasoning, where precision and factual correctness are essential for drawing valid inferences and supporting decision-making processes. Hallucinated outputs can lead to incorrect causal links, undermining the reliability of LLM-generated causal insights. Moreover, the lack of transparency in how LLMs generate such outputs further complicates efforts to identify and correct errors, posing challenges to researchers and practitioners alike.

Addressing these challenges requires strategies to ensure more controllable and precise causal data generation. High-quality causal data is essential not only for improving LLM accuracy but also for fostering trust in their outputs, particularly in real-world applications like healthcare, economics, and public policy. Future research should focus on robust methods to mitigate hallucinations and enhance the controllability of LLM-generated causal information, bridging the gap between language model capabilities and practical causality needs.

\section{Tasks \& Benchmarks Overview} 
\label{appendix:dataset_overview}
We first categorize the various end tasks into three groups: causal discovery, causal inference, and additional causal tasks.
Causal discovery involves identifying potential causal relationships between variables within large, often unannotated, datasets. This is crucial for understanding causal dynamics in complex systems. Causal inference refers to quantifying the impact of a treatment (or intervention) on an outcome, determining how changes in one variable causally influence another, typically using observational data.
Beyond these two primary tasks, there are additional causal tasks such as causal attribution, counterfactual reasoning, and causal explanation. These tasks extend the scope of causal analysis, often integrating causal reasoning with other reasoning tasks. 

We show the statistics of existing benchmarks for causal reasoning tasks (Table \ref{tab:dataset_statistics}) with those used in our experiments in bold. We also provide their inclusion of the three levels of causality: association, intervention, and counterfactual. 
From this data, we draw several key insights:
(1) \emph{Opportunity for Improvement}: There is a need for more datasets incorporating \textbf{intervention} and \textbf{counterfactual} reasoning to comprehensively evaluate causal reasoning.  
(2) \emph{Balanced Data}: Greater emphasis on \textbf{real-world} scenarios and underrepresented domains, such as health and social contexts, would enhance the robustness of causal benchmarks.  
(3) \emph{Scaling Challenges}: Although large datasets exist, developing datasets for complex causal tasks such as interventions remains difficult, limiting their representation. 
(4) \emph{Future Work}: Advancing causal reasoning in LLMs will require benchmarks that balance real and synthetic data, cover all three levels of causality, and span diverse domains.
In addition, we show examples from each dataset used in our experiments in Figure \ref{fig:prompt-instruction}.

\begin{figure}[t]
    \centering
    \includegraphics[width=\linewidth]{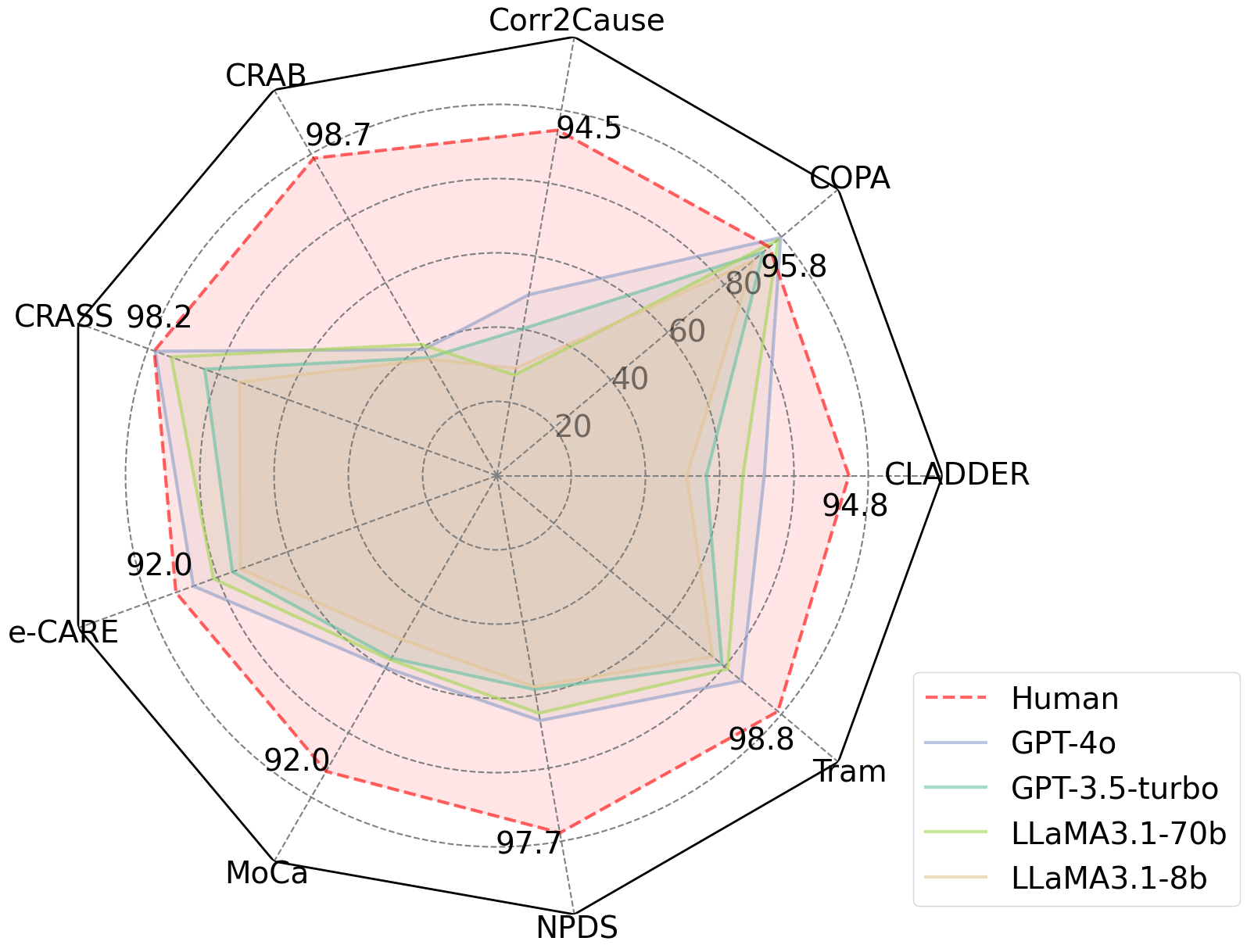}
    \caption{
     Overview of human and LLM performance on causal reasoning tasks, with annotated human scores.}
    \label{fig:radarchart}
\end{figure}

\section{Implementation Details}\label{appendix:implementation_details}

We begin by discussing the evaluation metrics used across different benchmarks. TRAM uses accuracy to assess causal reasoning tasks. NPDS evaluates performance with both F1 score and causal accuracy, while Corr2Cause employs F1 as the primary metric, alongside precision, recall, and accuracy. CRAB focuses on Pairwise and Graded Causality tasks, using Macro F1 for its 4-class Pairwise Causality Multiple Choice Question (MCQ) task. MOCA compares LLM responses with human participants using agreement (Agg), Area Under the Curve (AUC), Mean Absolute Error (MAE), and Cross-Entropy (CE), though we directly assess LLMs on causal tasks without relying on agreement metrics. CRASS uses top-1 accuracy, while e-CARE assesses MCQs with accuracy, average-BLEU (n=4), ROUGE-L, Perplexity, and Causal Explanation Quality (CEQ) for explanations. Finally, COPA uses accuracy as the primary metric.

Across these benchmarks, metrics such as F1, precision, recall, AUC, and BLEU assess different aspects of performance, including classification quality, alignment with human responses, and explanation quality. We select pass@1 accuracy as our primary metric because it is straightforward and widely used for evaluating multiple-choice and prediction-based causal tasks.

For evaluation, we randomly select 500 samples from each dataset's test set using a random seed of 42, or the maximum available size if fewer samples exist. Due to cost and rate constraints, models like GPT-4o and Claude-3.5 are tested on 100 samples. 
We utilize a unified prompt pattern (Figure \ref{fig:prompt-instruction}) comprising three parts: (1) \textbf{Identity Recognition}, such as "You are a highly intelligent question-answering bot with profound knowledge of causal inference"; (2) \textbf{Question-Context}, which describes the specific question to be addressed; and (3) \textbf{Answer Format Control}, which guides the model’s response. The latter instructs the model to either provide a direct answer with "Give me the answer directly" for Direct I/O or to begin with reasoning or evidence, concluding with the final result marked by "\#\#\#\#" for CoT reasoning, ensuring the output is easy to parse.
In the few-shot learning setup, we use two fixed examples for most datasets and one for the CRAB dataset, which has a longer context. Direct I/O prompts request direct answers, while CoT prompts involve generating CoT examples with GPT-4o (Figure \ref{fig:cot-example}), followed by manual validation for accuracy.
Pass @1 Accuracy is calculated using a string parsing function to extract answers from outputs tagged with "Answer" or "\#\#\#\#." Any model failing to follow the specified format is treated as a failure case.

We evaluate a diverse set of language models, including Mistral AI’s instructional models, Google’s Gemma models, Meta’s LLaMA 3 and 3.1 series, DeepSeek's V3 and R1 models, OpenAI’s GPT-3.5-turbo, GPT-4o, o1 and o3-mini, and Anthropic’s Claude-3.5, ensuring comprehensive coverage of various architectures and capabilities. 
For open-source models, we used the API provided by the Replicate and Together platform, while proprietary models were accessed through their respective company APIs.
The specific model versions used in our experiments are listed below:
\begin{itemize}
    \item mistralai/Mistral-7B-Instruct-v0.3
    \item mistralai/Mixtral-8x7B-Instruct-v0.1
    \item google/gemma-2-9b-it
    \item google/gemma-2-27b-it
    \item meta-llama/Meta-Llama-3-8B-Instruct
    \item meta-llama/Meta-Llama-3-70B-Instruct-Turbo
    \item meta-llama/Meta-Llama-3.1-8B-Instruct
    \item meta-llama/Meta-Llama-3.1-70B-Instruct-Turbo
    \item Deepseek-V3
    \item Deepseek-R1
    \item gpt-3.5-turbo-0125
    \item gpt-4o-2024-08-06
    \item claude-3-5-sonnet-20240620
    \item o1-2024-12-17
    \item o3-mini-2025-01-31
\end{itemize}

To assess human performance on each dataset, we first reviewed the original papers for the datasets and found that some \citep{du2022ecare, frohberg-binder-2022-crass, wang2023tram} already provided human evaluation results. For the remaining datasets, we asked three graduate students to solve 100 randomly selected problems from each dataset. Prior to the evaluation, the participants underwent training to develop a strong foundation in causal reasoning. They then solved the problems independently, and their average accuracy is calculated as the final result.

For our error analysis, we first categorize errors into four primary types. \textbf{Statistical errors} (SE) arise from issues in data collection, measurement, or analysis, distorting relationships between variables. Sub-classes include: \emph{confounding}, where an extraneous variable influences both independent and dependent variables, creating misleading associations; \emph{selection bias}, where the sample is unrepresentative of the target population, leading to biased estimates; and \emph{measurement error}, where inaccurate data collection results in flawed assessments. \textbf{Logical errors} (LE) involve flaws in reasoning, such as \emph{directionality errors} (misidentifying causal direction), \emph{inconsistency} (contradictory reasoning), and \emph{over-generalization} (drawing broad conclusions from specific cases without proper support). \textbf{Probabilistic errors} (PE) stem from incorrect handling of probabilities and statistical inferences, including \emph{Bayesian inference errors} (misapplication of priors or updates) and \emph{probability comparison errors} (misinterpreting conditional probabilities or neglecting base rates). Finally, \textbf{contextual errors} (CE) occur when subtle details, such as shifts in tone or minor actions, are overlooked, leading to flawed causal inferences.
We select three representative models (LLaMA3-8B, LLaMA3-70B, and GPT-4o) for error analysis. For each dataset, 100 incorrect test samples are randomly chosen and manually analyzed. The results and analysis are presented in Section \ref{analysis}.

\noindent {\textbf{Detailed Error Analysis}}
This appendix provides detailed examples and analysis of the error types discussed in Section \ref{analysis}. For each type, we present representative cases that illustrate common failure modes.

\noindent { \textbf{- Logical Errors - Case Study}}

\textbf{Context:} Billy and Suzy inherited an unusual type of hybrid car that has two special car batteries called Bartlett batteries. The car won't start unless it has at least one Bartlett battery. Since all the car needed was at least one Bartlett battery, they were able to start the car.

\textbf{Question:} Did Billy cause the car to start?

\textbf{Model's Answer:} A delightful question! Let's break down the stages: Billy brought a battery, Suzy also brought a battery, and either battery alone would start the car. Therefore, Billy's action was sufficient to cause the car to start. Answer: Yes

\textbf{Error Analysis:} The model exhibits several reasoning flaws. First, it conflates sufficiency with causation - while Billy's action was sufficient, it wasn't necessary. Second, it overlooks redundant causation by failing to recognize that Suzy's battery would have started the car regardless. Third, it relies on temporal sequence, implicitly assuming first action implies causation.

\noindent { \textbf{- Statistical Errors - Case Study}}

\textbf{Context:} A study tracked 1000 patients' recovery rates. In the data, 70\% of patients who exercised regularly recovered. Among those who didn't exercise, 60\% still recovered. The exercise group had more younger patients (avg age 45) than non-exercise group (avg age 65).

\textbf{Question:} Does exercise improve recovery rates?

\textbf{Model's Answer:} Let's analyze the data: The exercise group shows a 70\% recovery rate while the non-exercise group shows a 60\% recovery rate. There's a 10\% improvement in the exercise group. Answer: Yes, exercise clearly improves recovery rates.

\textbf{Error Analysis:} The model makes several statistical reasoning errors. It fails to account for selection bias in group composition and ignores age as a potential confounding variable. The model draws causal conclusions from raw correlations without controlling for population differences, missing the need for age-stratified analysis or matched comparisons.

\noindent { \textbf{- Probabilistic Errors - Case Study}}

\textbf{Context:} Imagine a self-contained, hypothetical world with only the following probabilities. Is the chance of kwox smaller when observing yupt?

\textbf{Model's Answer:} Let's calculate step by step: P(kwox|yupt) = 0.47 * 0.57 + 0.65 * 0.43 = 0.5141. Since P(kwox) $\approx$  0.5141 < 0.5, Answer: No.

\textbf{Error Analysis:} The model demonstrates several probabilistic reasoning failures. It incorrectly applies the conditional probability formula and shows confusion between joint and conditional probabilities. Most critically, it makes a comparison with the wrong baseline probability, leading to an invalid conclusion about relative likelihoods.

\noindent { \textbf{- Contextual Errors - Case Study}}

\textbf{Context:} In a small company with high employee turnover, all employees who left in the last quarter had completed their major projects. The company conducted exit interviews where employees cited various reasons for leaving including better opportunities, relocation, and personal reasons.

\textbf{Question:} Does project completion cause employee turnover?

\textbf{Model's Answer:} Looking at the data: All departing employees had completed their projects, and there is a perfect correlation between project completion and leaving. Answer: Yes, project completion appears to cause employee turnover.

\textbf{Error Analysis:} The model shows several contextual understanding failures. It ignores the broader context of employee career paths and job market dynamics, while failing to consider natural project completion as part of the normal work cycle. The model overlooks explicitly stated alternative reasons for leaving in the exit interviews. Furthermore, it misses crucial temporal aspects, namely that employees typically complete projects before leaving, regardless of their reason for departure.

\section{Comparison with Traditional Causal Reasoning Methods}
While our primary focus is on LLM-based approaches, we present comparison results with traditional causal reasoning methods using the Tübingen cause-effect pairs dataset, a well-established benchmark for causal direction identification, from \citet{kiciman2023causal}. As shown in Table~\ref{tab:tubingen_results}, traditional covariance-based methods like Slope~\citep{marx2017telling}, bQCD~\citep{pmlr-v119-tagasovska20a}, and PNL-MLP~\citep{zhang2012identifiability} achieve accuracies ranging from 0.68 to 0.75, while Mosaic~\citep{wu2020mosaic} reaches 0.83. In comparison, \citet{kiciman2023causal}'s prompting strategies with LLMs demonstrate progressively improving performance: from 0.81 with basic GPT-3.5-turbo (without any special prompting), improving to 0.86 with causal agent prompting (using the system message "You are a helpful assistant for causal reasoning"), and further reaching 0.89 with single prompt engineering (reformulating the task as a direct question about causal direction with step-by-step reasoning requirements). Their experiments with GPT-4 using the same single prompt strategy ultimately achieved 0.96 accuracy. These results suggest that well-designed prompting techniques can effectively leverage LLMs' inherent understanding of causal relationships, outperforming traditional statistical approaches on this benchmark dataset.

\begin{table}[t]
\centering
\small
\begin{tabular}{lr}
\toprule
Model & Acc. \\
\midrule
Slope~\citep{marx2017telling} & 0.75 \\
bQCD~\citep{pmlr-v119-tagasovska20a} & 0.68 \\
PNL-MLP~\citep{zhang2012identifiability} & 0.75 \\
Mosaic~\citep{wu2020mosaic} & 0.83 \\
\midrule
GPT-3.5-turbo & 0.81 \\
GPT-3.5-turbo (causal agent)  & 0.86 \\
GPT-3.5-turbo (single prompt) & 0.89 \\
GPT-4 (single prompt)  & \textbf{0.96} \\ 
\bottomrule
\end{tabular}
\caption{Performance comparison between traditional causal discovery methods and LLM-based approaches on the Tübingen cause-effect pairs benchmark. Traditional methods rely on statistical analysis of observational data, while LLM-based approaches leverage natural language understanding.}
\label{tab:tubingen_results}
\end{table}

\clearpage

\begin{figure*}[h]
    \centering
    \footnotesize
    \demonstrationfigure{\textsc{Cladder}}{
        Question: \{Imagine a self-contained, hypothetical world with only the following conditions, and without any unmentioned factors or causal relationships: Full moon has a direct effect on sprinkler and weather. Weather has a direct effect on ground. Sprinkler has a direct effect on ground. For situations where there is no full moon, the probability of wet ground is 38\%. For situations where there is a full moon, the probability of wet ground is 43\%. Does full moon negatively affect ground through weather and sprinkler?\} \newline
        [Direct I/O] Give me the anwser directly. Answer: Yes or Answer: No. Do not use any other format. \newline
        [CoT] Begin your response with reasoning or evidence to suport your explanation, then return me the final result marked by '\#\#\#\#'.
        The answer format is \#\#\#\#YES or \#\#\#\#NO.
    }
    \demonstrationfigure{\textsc{COPA}}{
        Premise: \{The man's voice sounded hoarse.\} \newline
        Question: What was the \{ What was the cause?\}? \newline
        A. \{He had a cold.\} \newline
        B. \{He quit smoking.\} \newline
         [Direct I/O] Give me the anwser directly. Answer: A or Answer: B. Do not use any other format. \newline
        [CoT] Begin your response with reasoning or evidence to suport your explanation, then return me the final result marked by '\#\#\#\#'. The answer format is \#\#\#\#A or \#\#\#\#B.
    }
    \demonstrationfigure{\textsc{CRAB}}{
        Context: \newline
        Article A: \{International pressure may be the only way to stop the Brazilian government from taking a \u201csuicide\u201d path in the Amazon, one of the country\u2019s most respected scientists has said, as the world\u2019s biggest rainforest continues...\} \newline
        Article B: \{Hollywood star Leonardo DiCaprio has announced a \$5 million pledge to help reduce the number of fires in the Amazon rainforest. It comes amid global outcry over fires in the Brazilian Amazon, with celebrities calling for action and pledging to donate to efforts in order to tackle the issue.DiCaprio announced on Sunday that Earth Alliance, an organization dedicated to environmental philanthropy that he co-chairs, was pledging \$5 million...\} \newline
        Event 1: \{The EU has recently signed a trade deal with South America.\} \newline
        Event 2: \{Leonardo DiCaprio pledges \$5 million to help reduce the number of fires in the Amazon rainforest.\} \newline
        How much did event 1 cause event 2 to happen? \newline
        [A] High causality: Event 1 is definitely responsible for Event 2. \newline
        [B] Medium causality: Event 1 might have been responsible for Event 2. \newline
        [C] Low causality: The context gives a little indication that there is a connection between the two events, but background info might suggest a low causal connection. \newline
        [D] No causality: Events are somehow related but definitely NOT causally related. \newline
        [Direct I/O] Answer: A, or Answer: B, or Answer: C, or Answer: D. Do not use any other format. \newline
        [CoT] Begin your response with reasoning or evidence to suport your explanation, then return me the final result marked by '\#\#\#\#'. The answer format is \#\#\#\#A or \#\#\#\#B or \#\#\#\#C or \#\#\#\#D.
    }
    \demonstrationfigure{\textsc{CRASS}}{
        Question: \{A woman opens a treasure chest. What would have happened if the woman had not opened the treasure chest?\} \newline
        Options:  \newline A)\{The treasure chest would have been open.\}, \newline B)\{That is not possible.\}, \newline C)\{The treasure chest would have remained closed.\}, \newline D)\{I don't know.\} \newline
        [Direct I/O] Answer: A, or Answer: B, or Answer: C, or Answer: D. Do not use any other format. \newline
        [CoT] Begin your response with reasoning or evidence to suport your explanation, then return me the final result marked by '\#\#\#\#'. The answer format is [A] or [B] or [C] or [D].
    }
    \demonstrationfigure{\textsc{E-CARE}}{
        Premise:\{Tom wants to go to space.\} \newline
        A. \{He aspired to be an astronaut.\} \newline
        B. \{He aspired to be a calligrapher.\} \newline
        Which of the following is more likely to be true? \newline
        [Direct I/O] Answer: A or Answer: B. Do not use any other format. \newline
        [CoT] Begin your response with reasoning or evidence to suport your explanation, then return me the final result marked by '\#\#\#\#'. The answer format is \#\#\#\#A or \#\#\#\#B.
    }
    
\end{figure*}

\begin{figure*}[h]
    \centering
    \footnotesize
    \demonstrationfigure{\textsc{MOCA}}{
        Story: \{Sara has an electric guitar that she plays through an amplifier. The guitar will play through the amplifier if either the amplifier is set to channel A or the power mixer is turned on. Today, the amplifier is set to channel A. Sara checks to see if the power mixer is on. She sees that the power mixer is turned on, and she does not turn off the power mixer or adjust it at all. Because the guitar will play through the amplifier if either the amplifier is set to channel A or the power mixer is turned on, Sara's guitar plays through the amplifier.\} \newline
        Question: \{Did the guitar play through the amplifier because Sara did not turn off the power mixer?\} \newline
        [Direct I/O] Answer: Yes or Answer: No. Do not use any other format. \newline
        [CoT] Begin your response with reasoning or evidence to suport your explanation, then return me the final result marked by '\#\#\#\#'. The answer format is \#\#\#\#YES or \#\#\#\#NO.
    }
    \demonstrationfigure{\textsc{NPDS}}{
    Question: \{L Achillodyni causes R Br\u00f6stbesv\u00e4r. (``R'' and ``L'' refer to the right and left sides of the body, respectively). Answer with true or false.\} \newline
    [Direct I/O] Answer: True or Answer: False. Do not use any other format. \newline
    [CoT] Begin your response with reasoning or evidence to support your explanation, then return me the final result marked by '\#\#\#\#'. The answer format is \#\#\#\#TRUE or \#\#\#\#FALSE.
    }
    \demonstrationfigure{\textsc{TRAM}}{
        Premise:\{As evening approaches, households engage in their routine culinary activities.\} \newline
        Question:\{What's the more plausible RESULT?\} \newline
        A. \{People celebrated the start of the new year.\} \newline
        B. \{People prepared their dinner.\} \newline
        [Direct I/O] Answer: A or Answer: B. Do not use any other format. \newline
        [CoT] Begin your response with reasoning or evidence to suport your explanation, then return me the final result marked by '\#\#\#\#'. The answer format is \#\#\#\#A or \#\#\#\#B.
    }
    \demonstrationfigure{\textsc{CORR2CAUSE}}{
        Question: \{Suppose there is a closed system of 5 variables, A, B, C, D and E. All the statistical relations among these 5 variables are as follows: A correlates with B. A correlates with C. A correlates with D. A correlates with E. B correlates with C. B correlates with D. B correlates with E. C correlates with D. C correlates with E. D correlates with E. However, C and E are independent given A and B. C and E are independent given A, B and D. D and E are independent given A and B. D and E are independent given A, B and C. Determine the truth value the following statement: A influences B through some mediator(s).\} \newline
        Determine the truth value the following statement: \{A influences B through some mediator(s).\} \newline
        [Direct I/O] Answer: neutral or Answer: contradiction or Answer: entailment. Do not use any other format. \newline
        [CoT] Begin your response with reasoning or evidence to suport your explanation, then return me the final result marked by '\#\#\#\#'. The answer format is \#\#\#\#NEUTRAL or \#\#\#\#CONTRADICTION or \#\#\#\#ENTAILMENT.
        }
\caption{Prompts for evaluating LLMs on different casual reasoning tasks.}
\label{fig:prompt-instruction}
\end{figure*}

\begin{figure*}[h]
\demonstrationfigure{\textsc{COPA COT Example}}{
        Premise: The man's voice sounded hoarse.
        
        Question: What was the cause?
        
        A. He had a cold.
        
        B. He quit smoking.\newline
        
         Begin your response with reasoning or evidence to support your explanation, then return me the final result marked by '\#\#\#\#'.\newline
        
        A hoarse voice is often a symptom of a cold, as it can result from inflammation of the vocal cords caused by infection. Quitting smoking, while beneficial for overall health, generally leads to improved vocal quality over time rather than immediate hoarseness.\newline

       \#\#\#\# A
}

\demonstrationfigure{\textsc{NPDS COT Example}}{
    Question: L Achillodyni causes R Br\"ostbesv\"ar. 'R' and 'L' refer to the right and left sides of the body, respectively. 
    
    Answer with true or false.\newline
    
    Begin your response with reasoning or evidence to support your explanation, then return me the final result marked by '\#\#\#\#'.\newline

    Achillodynia refers to pain in the Achilles tendon, while br\"ostbesv\"ar refers to chest pain or chest-related issues. These are two distinct anatomical regions: the Achilles tendon is located in the lower leg, and chest pain pertains to the upper torso. There is no direct physiological connection between Achilles tendon issues and chest pain. Therefore, the claim that left-sided Achillodynia causes right-sided Br\"ostbesv\"ar lacks anatomical or medical evidence.\newline

    \#\#\#\# False
}

\demonstrationfigure{\textsc{TRAM COT Example}}{
    As evening approaches, households engage in their routine culinary activities.\newline
    
    What's the more plausible RESULT?\newline
    A. People celebrated the start of the new year.\newline
    B. People prepared their dinner.\newline
    
    Begin your response with reasoning or evidence to support your explanation, then return me the final result marked by '\#\#\#\#'.\newline

    As evening approaches, households typically engage in their routine activities, which often include preparing meals. This situation aligns more naturally with the daily routine of preparing dinner rather than an annual event like celebrating the new year. Celebrating the new year is a rare and specific occasion, while preparing dinner is a common, everyday activity.
    The evidence suggests that dinner preparation is a more plausible result in this context.\newline

    \#\#\#\# B
}

\demonstrationfigure{\textsc{e-CARE COT Example}}{
    Tom wants to go to space.\newline
    A. He aspired to be an astronaut.\newline
    B. He aspired to be a calligrapher.\newline
    Which of the following is more likely to be true?\newline
    
    Begin your response with reasoning or evidence to support your explanation, then return me the final result marked by '\#\#\#\#'.\newline

    Since Tom wants to go to space, the likelihood of this aspiration aligning with a related field or interest is greater. Aspiring to be an astronaut is directly connected to space exploration, and people who want to go to space typically have a background or interest in relevant fields such as science, engineering, or aeronautics.\newline

    Aspiring to be a calligrapher, while a valuable skill, does not naturally correlate with a desire to go to space, as it is unrelated to space exploration or the skill sets typically required for such an endeavor.\newline

    \#\#\#\# A
}
\end{figure*}

\begin{figure*}[h]
\demonstrationfigure{\textsc{CLADDER COT Example}}{
    Question: Imagine a self-contained, hypothetical world with only the following conditions, and without any unmentioned factors or causal relationships: Full moon has a direct effect on sprinkler and weather. Weather has a direct effect on ground. Sprinkler has a direct effect on ground. For situations where there is no full moon, the probability of wet ground is 38\%. For situations where there is a full moon, the probability of wet ground is 43\%. Does full moon negatively affect ground through weather and sprinkler?\newline 
    
    Begin your response with reasoning or evidence to support your explanation, then return me the final result marked by '\#\#\#\#'.\newline

    To determine whether the full moon negatively affects the ground through weather and sprinkler, we need to consider the chain of causal relationships in this hypothetical world. The full moon has a direct effect on both the sprinkler and the weather, which in turn directly affect the ground.\newline
    Here are the key details:\newline
    - When there is **no full moon**, the probability of wet ground is 38\%.\newline
    - When there **is a full moon**, the probability of wet ground is 43\%.\newline

    If the full moon negatively affected the ground through the weather and sprinkler, we would expect the probability of wet ground to decrease in the presence of a full moon. However, the probability of wet ground increases (from 38\% to 43\%) when there is a full moon.\newline

    Thus, based on this evidence, the full moon does not negatively affect the ground through the weather and sprinkler. Instead, it seems to have a positive or neutral effect.\newline

    \#\#\#\# No
}

\demonstrationfigure{\textsc{CRASS COT Example}}{
    Question: A girl reveals a thing. What would have happened if the girl had concealed the thing?\newline 
    A) The thing would have been revealed.\newline 
    B) That is not possible.\newline 
    C) Everything would have been fine.\newline 
    D) The thing would have not been visible.\newline 
    
    Begin your response with reasoning or evidence to support your explanation, then return me the final result marked by '\#\#\#\#'.\newline

    To answer this question from a causal inference perspective, we must examine the hypothetical scenario where the girl did not reveal the thing (i.e., she concealed it). The question is asking us to compare the actual world (where the thing was revealed) to the counterfactual world (where the thing would have been concealed).\newline
    - \textbf{Option A}: "The thing would have been revealed" directly contradicts the hypothetical situation because the girl concealing the thing would prevent it from being revealed.\newline
    - \textbf{Option B}: "That is not possible" might refer to the idea that the girl cannot conceal the thing in some absolute sense, but no evidence suggests that such a scenario is impossible. Thus, this is an unlikely answer.\newline
    - \textbf{Option C}: "Everything would have been fine" introduces an assumption about the broader consequences of the concealment, which is speculative and not necessarily implied by the question.\newline
    - \textbf{Option D}: "The thing would have not been visible" aligns with the hypothetical situation where concealment prevents the thing from being revealed. This directly follows from the premise of the question.\newline

    Based on this reasoning, the most plausible answer is that if the girl had concealed the thing, it would not have been visible.\newline

    \#\#\#\# D
}
\end{figure*}

\begin{figure*}[h]
\demonstrationfigure{\textsc{MoCa COT Example}}{
    Story: Sara has an electric guitar that she plays through an amplifier. The guitar will play through the amplifier if either the amplifier is set to channel A or the power mixer is turned on. Today, the amplifier is set to channel A. Sara checks to see if the power mixer is on. She sees that the power mixer is turned on, and she does not turn off the power mixer or adjust it at all. Because the guitar will play through the amplifier if either the amplifier is set to channel A or the power mixer is turned on, Sara's guitar plays through the amplifier.\newline 
    
    Question: Did the guitar play through the amplifier because Sara did not turn off the power mixer?\newline 
    
    Begin your response with reasoning or evidence to support your explanation, then return me the final result marked by '\#\#\#\#'.\newline 

    The guitar will play through the amplifier if either the amplifier is set to channel A or the power mixer is turned on. In this case, we know that the amplifier is already set to channel A, which alone is sufficient for the guitar to play through the amplifier, regardless of the power mixer's status.\newline

    Even though Sara did not turn off the power mixer, the amplifier being set to channel A was already enough for the guitar to work. Therefore, the guitar would have played through the amplifier even if the power mixer had been off or turned on. Sara's action of not turning off the power mixer had no effect on the outcome.\newline

    \#\#\#\# No
}

\demonstrationfigure{\textsc{Corr2cause COT Example}}{
    Question: Suppose there is a closed system of 5 variables, A, B, C, D, and E. All the statistical relations among these 5 variables are as follows: A correlates with B. A correlates with C. A correlates with D. A correlates with E. B correlates with C. B correlates with D. B correlates with E. C correlates with D. C correlates with E. D correlates with E. However, C and E are independent given A and B. C and E are independent given A, B, and D. D and E are independent given A and B. D and E are independent given A, B, and C.\newline
    
    Determine the truth value of the following statement: A influences B through some mediator(s).\newline
    
    Begin your response with reasoning or evidence to support your explanation, then return me the final result marked by '\#\#\#\#'.\newline

    The statement asks whether A influences B through some mediator(s). Let's reason through this based on the given information about correlations and conditional independencies.
    
    1. Correlation Relationships: A correlates with B, C, D, and E. Similarly, B correlates with C, D, and E. This suggests that A and B are connected, but it doesn't yet clarify whether the relationship is direct or mediated.
    
    2. Conditional Independencies:
    
    - C and E are independent given A and B, and also independent given A, B, and D. This tells us that once we know A and B, C and E do not provide additional information about each other.
    
    - D and E are independent given A and B, and also independent given A, B, and C. This implies that once A and B are known, D and E do not add any further influence over each other.
    
    These independencies suggest that the relationships between variables are conditional, possibly implying a complex interaction but without directly stating that a mediator is necessary between A and B.
    
    3. No Direct Evidence of Mediation: The problem specifies conditional independencies between some variables given others, but there is no clear indication of a mediating variable between A and B. The provided conditions do not imply a mediation relationship but rather highlight correlations and conditional independencies.\newline

    Given that no mediating variable is directly implicated between A and B from the given information, the truth value of the statement "A influences B through some mediator(s)" cannot be definitively concluded as true (entailment) or false (contradiction). Instead, the information is insufficient to prove or disprove mediation.\newline

    \#\#\#\# neutral
}
\end{figure*}

\begin{figure*}[h]
    \centering
    \footnotesize
\demonstrationfigure{\textsc{CRAB COT Example}}{
    Context:\newline
    Article A: \newline
    International pressure may be the only way to stop the Brazilian government from taking a “suicide” path in the Amazon, one of the country’s most respected scientists has said, as the world’s biggest rainforest continues to be ravaged by thousands of deliberate fires.\newline 
    The large number of conflagrations – set illegally to clear and prepare land for crops, cattle and property speculation – has prompted the state of Amazonas to declare an emergency, created giant smoke clouds that have drifted hundreds of miles, and sparked international concerns about the destruction of an essential carbon sink.\newline 
    “Our house is burning,” tweeted the French president, Emmanuel Macron, who called for emergency talks on the subject at this week’s G7 summit. But the response to the crisis has been mixed: while Norway and Germany have halted donations to the Brazilian government’s Amazon fund, the EU has recently signed a trade deal with South America, and the UK spent this week focusing on post-Brexit business with Brazil.\newline 
    Scientists say the ongoing destruction will have dire consequences for Brazil and the world.\newline

    Article B: \newline
    Hollywood star Leonardo DiCaprio has announced a \$5 million pledge to help reduce the number of fires in the Amazon rainforest.\newline 
    It comes amid global outcry over fires in the Brazilian Amazon, with celebrities calling for action and pledging to donate to efforts in order to tackle the issue.\newline 
    DiCaprio announced on Sunday that Earth Alliance, an organization dedicated to environmental philanthropy that he co-chairs, was pledging \$5 million in funding to protect the Amazon.\newline 
    “The governments around the world, including Brazil, need to work together to make sure this doesn’t continue,” he said.\newline 

    Event 1: The EU has recently signed a trade deal with South America.\newline 
    Event 2: Leonardo DiCaprio pledges \$5 million to help reduce the number of fires in the Amazon rainforest.\newline

    How much did event 1 cause event 2 to happen?\newline
    [A] High causality: Event 1 is definitely responsible for Event 2.\newline 
    [B] Medium causality: Event 1 might have been responsible for Event 2.\newline 
    [C] Low causality: The context gives a little indication that there is a connection between the two events, but background info might suggest a low causal connection.\newline 
    [D] No causality: Events are somehow related but definitely NOT causally related.\newline
    
    Begin your response with reasoning or evidence to support your explanation, then return me the final result marked by '\#\#\#\#'.\newline

    Event 1, the EU signing a trade deal with South America, and Event 2, Leonardo DiCaprio pledging \$5 million to help reduce fires in the Amazon, are both related to the broader context of the Amazon rainforest fires and global reactions. However, the two events appear to stem from different sources of influence and motivation.\newline

    Event 1 focuses on trade and economic relations between the EU and South America, whereas Event 2 involves an individual actor (DiCaprio) taking philanthropic action. The trade deal seems to be driven by economic priorities, while DiCaprio’s pledge is motivated by environmental concerns and public outcry. There is no clear evidence or connection that suggests the trade deal directly caused DiCaprio to pledge his donation. The timing of both events might be coincidental, as they both respond to the ongoing Amazon crisis, but they do not seem to be causally related.\newline

    Thus, the causality between the two events is likely very low, if not non-existent.\newline

    \#\#\#\# D

}

\caption{The CoT examples generated by GPT-4o.}
\label{fig:cot-example}
\end{figure*}

\begin{figure*}[h]
    \centering
    \footnotesize
\demonstrationfigure{\textsc{Casual COT Example}}
{
Question:Imagine a self-contained, hypothetical world with only the following conditions, and without any unmentioned
factors or causal relationships: Full moon has a direct effect on sprinkler and weather. Weather has a direct effect
on ground. Sprinkler has a direct effect on ground. For situations where there is no full moon, the probability of
wet ground is 38\%. For situations where there is a full moon, the probability of wet ground is 43\%. Does full moon
negatively affect ground through weather and sprinkler?

Guidance: Address the question by following the steps below: 

Step 1) Extract the causal graph: Identify the causal graph that depicts the relationships in the scenario. The diagram should simply consist of edges denoted in "var1-> var2" format, separated by commas.

Step 2) Determine the query type: Identify the type of query implied by the main question. Choices include "marginal probability", "conditional probability", "explaining away effect", "backdoor ad justment set", "average treatment effect", "collider bias", "normal counterfactual question", "average treatment effect on treated", "natural direct effect" or "natural indirect effect". Your answer should only be a term from the list above, enclosed in quotation marks.

Step 3) Formalize the query: Translate the query into its formal mathematical expression based on its type, utilizing the "do(·)" notation or counterfactual notations as needed. 

Step 4) Gather all relevant data: Extract all the available data. Your answer should contain nothing but marginal probabilities and conditional probabilities in the form "P(...)=..." or "P(...|...)=...", each probability being separated by a semicolon. Stick to the previously mentioned denotations for the variables.

Step 5) Deduce the estimand using causal inference: Given all the information above, deduce the estimand using skills such as do-calculus, counterfactual prediction, and the basics of probabilities. Answer step by step.

Step 6) Calculate the estimand: Insert the relevant data in Step 4 into the estimand, perform basic arithmetic calculations, and derive the final answer. There is an identifiable answer. Answer step by step.

Begin your response with reasoning or evidence to support your explanation, then return me the final result marked by

\#\#\#\#.

Let's solve the question step-by-step as per the provided guidance:

\#\#\# Step 1: Extract the causal graph
The relationships described are:

- Full moon affects sprinkler and weather.
- Weather affects ground.
- Sprinkler affects ground.

The causal graph is:

```
FullMoon -> Weather, FullMoon -> Sprinkler, Weather -> Ground, Sprinkler -> Ground
```

\#\#\# Step 2: Determine the query type
The query asks whether the **full moon negatively affects the ground through weather and sprinkler**. This involves understanding the **total effect** of Full Moon on the probability of wet ground, considering direct and indirect causal pathways.

The correct term is **"natural indirect effect"**.

\#\#\# Step 3: Formalize the query
We want to assess the effect of **Full Moon** (F) on **Ground** (G) mediated through **Weather** (W) and **Sprinkler** (S). This can be expressed as:

\[
\text{NIE} = E[G \mid \text{do}(F = 1)] - E[G \mid \text{do}(F = 0)]
\]

Where:
- \( G \) = Ground
- \( F = 1 \) (Full Moon present), \( F = 0 \) (Full Moon absent)
- "do" notation removes any confounding to isolate causal effects.

\#\#\# Step 4: Gather all relevant data
From the problem, we are given:

1. \( P(G = 1 \mid F = 0) = 0.38 \) (No full moon → wet ground).
2. \( P(G = 1 \mid F = 1) = 0.43 \) (Full moon → wet ground).

\#\#\# Step 5: Deduce the estimand using causal inference
To determine whether **Full Moon negatively affects Ground through Weather and Sprinkler**, we compare the probabilities \( P(G = 1 \mid F = 1) \) and \( P(G = 1 \mid F = 0) \). Specifically:

- If \( P(G = 1 \mid F = 1) > P(G = 1 \mid F = 0) \), then the Full Moon **positively** affects Ground.
- If \( P(G = 1 \mid F = 1) < P(G = 1 \mid F = 0) \), then the Full Moon **negatively** affects Ground.

Here, the difference:

\[
\Delta = P(G = 1 \mid F = 1) - P(G = 1 \mid F = 0)
\]

\#\#\# Step 6: Calculate the estimand
Substitute the values given:

\[
\Delta = P(G = 1 \mid F = 1) - P(G = 1 \mid F = 0)
\]
\[
\Delta = 0.43 - 0.38
\]
\[
\Delta = 0.05
\]

Since \( \Delta > 0 \), the Full Moon **positively** affects the probability of wet ground through Weather and Sprinkler. Therefore, there is **no negative effect**.

\#\#\#\# Final Result
\#\#\#\# No

}
\caption{The causalCoT prompt used in our evaluation.}
\label{fig:causalcot-example}
\end{figure*}

\begin{figure*}[h]
    \centering
    \footnotesize
\demonstrationfigure{\textsc{Fuctiona Call Example}}
{
Question:\\
Imagine a self-contained, hypothetical world with only the following conditions, and without any unmentioned factors or causal relationships: 
Smoking has a direct effect on effort and freckles. Effort has a direct effect on freckles. For nonsmokers who are lazy, the probability of freckles is $92\%$. For nonsmokers who are hard-working, the probability of freckles is $50\%$. For smokers who are lazy, the probability of freckles is $80\%$. For smokers who are hard-working, the probability of freckles is $36\%$. 
For nonsmokers, the probability of being hard-working is $17\%$. For smokers, the probability of being hard-working is $74\%$. If we disregard the mediation effect through effort, would smoking negatively affect freckles?
Begin your response with reasoning or evidence to support your explanation, then return the final result marked by \texttt{\#\#\#\#}. The answer format is \texttt{\#\#\#\# Yes} or \texttt{\#\#\#\# No}.
\\
\textbf{Code interpreter request:} 
Given probabilities:\\
p\_f\_nonsmoker\_lazy = 0.92\\
p\_f\_nonsmoker\_hardworking = 0.50\\
p\_f\_smoker\_lazy = 0.80\\
p\_f\_smoker\_hardworking = 0.36\\
Given probability of being hardworking:\\
p\_hardworking\_nonsmoker = 0.17\\
p\_hardworking\_smoker = 0.74\\
Derive probabilities of being lazy:\\
p\_lazy\_nonsmoker = 1 - p\_hardworking\_nonsmoker\\
p\_lazy\_smoker = 1 - p\_hardworking\_smoker\\
Calculate weighted probabilities:\\
p\_freckles\_nonsmoker = (p\_f\_nonsmoker\_lazy * p\_lazy\_nonsmoker) +
(p\_f\_nonsmoker\_hardworking * p\_hardworking\_nonsmoker)\\
p\_freckles\_smoker = (p\_f\_smoker\_lazy * p\_lazy\_smoker) +
(p\_f\_smoker\_hardworking * p\_hardworking\_smoker)\\
p\_freckles\_nonsmoker, p\_freckles\_smoker\\
\textbf{Code interpreter response:} Code executed successfully.\\
\textbf{Code interpreter request:} \\
\# Output the results\\
p\_freckles\_nonsmoker, p\_freckles\_smoker\\
\textbf{Code interpreter response:} $(0.8486, 0.4744)$\\
\textbf{Final response:} \\
The computed probabilities of having freckles, when disregarding the mediation effect through effort, are:\\
\[ P(\text{freckles } | \text{ nonsmoker}) = 84.86\% \]
\[ P(\text{freckles } | \text{ smoker}) = 47.44\% \]
\textbf{Conclusion}
By disregarding the mediation effect through effort, smoking appears to have a negative effect on the probability of developing freckles. Smokers have a lower probability of having freckles compared to nonsmokers.\\
\texttt{\#\#\#\# No}
}
\caption{A Function Call example from CLADDER dataset evaluation.}
\label{fig:causalcot-example}
\end{figure*}

\end{document}